\documentclass{article}

\PassOptionsToPackage{numbers, compress}{natbib}
 \usepackage[preprint]{neurips_2026}

\usepackage{amsmath,amsfonts,bm}

\def\eqref#1{equation~\ref{#1}}

\def\1{\bm{1}}

\def\vtheta{{\bm{\theta}}}
\def\vtau{{\bm{\tau}}}

\def\vg{{\bm{g}}}

\def\vx{{\bm{x}}}
\def\vy{{\bm{y}}}
\def\vz{{\bm{z}}}

\def\mG{{\bm{G}}}

\def\mI{{\bm{I}}}

\def\mR{{\bm{R}}}

\def\mX{{\bm{X}}}

\DeclareMathAlphabet{\mathsfit}{\encodingdefault}{\sfdefault}{m}{sl}
\SetMathAlphabet{\mathsfit}{bold}{\encodingdefault}{\sfdefault}{bx}{n}

\usepackage[utf8]{inputenc} 
\usepackage[T1]{fontenc}    
\usepackage[hidelinks]{hyperref}
\usepackage{url}            
\usepackage{booktabs}       
\usepackage{amsfonts}       
\usepackage{nicefrac}       
\usepackage{microtype}      
\usepackage{xcolor}         
\usepackage{graphicx}
\usepackage{makecell}
\usepackage{amsmath}
\usepackage{wrapfig}
\usepackage{algorithm}
\usepackage{algorithmic}
\usepackage{amsmath}
\usepackage{amssymb}
\usepackage{caption}
\usepackage[table]{xcolor}

\newcommand{\mymethod}{{BiCo}}
\newcommand{\baik}[1]{\textcolor{red}{#1}}

\title{Bilinear Coordinate Alignment \\for Training-Free Task-Vector Transfer}

\author{%
Jungyong Son\textsuperscript{1}, Jinwook Jung\textsuperscript{1}, Minhee Park\textsuperscript{1}, Sungyong Baik\textsuperscript{1,2}\thanks{Corresponding author.}\\
\textsuperscript{1} Dept. of Artificial Intelligence, \textsuperscript{2} Dept. of Data Science \\
Hanyang University\\
  \texttt{\{jungyongs,jjw970517,himinsunsine,dsybaik\}@hanyang.ac.kr} \\
}

\begin{document}

\maketitle
\begin{abstract}
Fine-tuning large-scale pre-trained models is a recent prevalent paradigm for adapting general representations to specialized tasks.
However, when a new version of a pre-trained model becomes available, expertise acquired through fine-tuning cannot be directly reused because it is tied to the parameterization of the original model, requiring another costly fine-tuning.
To address this inefficiency, recent work uses task vectors—defined as the parameter difference between a fine-tuned model and its base model—to transfer expertise across models.
While existing methods bridge disparate models by matching activations or gradients, a significant performance gap remains relative to direct fine-tuning, suggesting that these partial correspondences are insufficient.
In this work, instead of viewing a task vector merely as a parameter offset, we revisit the formation of task vectors and show that they can be derived as accumulated bilinear interactions between input-side activations and output-side gradients.
Motivated by this observation, we formulate task-vector transfer as a dual-space alignment problem and propose \textbf{\mymethod{}}, a training-free framework for transferring task vectors through \textbf{Bi}linear \textbf{Co}ordinate alignment.
\mymethod{} estimates orthogonal Procrustes mappings in both spaces using a single forward--backward pass on a small calibration set, without any parameter update.
Across extensive computer vision and natural language processing benchmarks, \mymethod{} consistently outperforms existing transfer methods across models that differ in width, depth, and pre-training configuration.
\end{abstract}

\section{Introduction}
\label{sec:introduction}
Modern machine learning has increasingly adopted a paradigm in which large-scale pre-trained models~\citep{achiam2023gpt,touvron2023llama,transformer} are adapted to downstream tasks through task-specific fine-tuning~\citep{brown2020language,wei2022finetuned,wortsman2022robust}. 
While this approach is highly effective, it becomes inefficient as pre-trained model families are continually expanded with new variants that follow similar architectural designs but differ in model scale, pre-training data, or training recipes~\citep{clip,gadre2023datacomp}.
Since fine-tuned knowledge is encoded in the parameter space of the specific pre-trained model on which it was learned, it cannot be directly transferred to a different model variant.
Consequently, each newly released pre-trained model often requires separate fine-tuning to acquire similar task-specific capabilities, leading to repeated computational overhead and increased storage demands.
This work addresses the challenge of transferring task-specific knowledge across pre-trained model variants without requiring additional fine-tuning.

To represent and manipulate this fine-tuned expertise, Task Arithmetic~\citep{ilharco2023editing} introduces the concept of task vectors, defining them as the parameter difference between a fine-tuned model and its base model (i.e., $\boldsymbol{\tau} = \vtheta_{\mathrm{ft}} - \vtheta_{\mathrm{pre}}$). 
This formulation has led to a growing line of research on task-vector transfer~\citep{update,gradfix,theseus}: given a task-vector obtained by fine-tuning a source pre-trained model, the goal is to transfer this vector to a separate target pre-trained model so that the target acquires the corresponding task-specific capability without undergoing full fine-tuning.
One prominent direction exploits permutation symmetries to align the weight spaces of different models~\citep{gitrebasin,update}. 
However, such parameter-level alignment typically relies on strong architectural compatibility and matching parameter dimensions, limiting its applicability to models that differ in scale or width. 
To overcome these limitations, recent studies have explored using task-specific signals, such as gradients or activations, to transform task vectors across disparate models~\citep{theseus,gradfix}.
GradFix~\cite{gradfix} uses target-model gradients to mask task vectors
according to the local loss landscape, while THESEUS~\cite{theseus} transfers task vectors
by matching intermediate activation spaces.
While these approaches provide valuable partial correspondences, a significant performance gap remains relative to direct fine-tuning.

In this work, we first show that a task vector can be derived as  the accumulation of the bilinear interactions coupling input-side activations with output-side gradients.
Building on this insight, we argue that treating activations and gradients as separate, independent proxies for task knowledge fails to capture the underlying mechanics of the parameter update.
Rather than naively combining these disjoint alignment strategies, we formulate task-vector transfer as a holistic, coupled dual-space alignment problem.

To this end, we propose \mymethod{}, a training-free framework that leverages \textbf{Bi}linear \textbf{Co}ordinate alignment to transfer task-specific expertise across pre-trained models.
By applying orthogonal Procrustes alignment~\citep{schonemann1966generalized} to both spaces, \mymethod{} transforms the source task vector into a target-compatible update while preserving the geometry of the aligned feature and gradient spaces.
Importantly, the required alignments are estimated from a small calibration set using a single forward--backward pass through frozen pre-trained models, requiring neither parameter updates nor task-specific hyperparameter tuning.
We demonstrate the effectiveness of \mymethod{} across a wide range of benchmarks in computer vision and natural language processing, showing that it enables robust knowledge transfer even across models with architectural and pre-training variations.
Our contributions are summarized as follows:
\begin{itemize}
\item We revisit the mechanistic formation of task vectors and formulate their transfer as a dual-space coordinate alignment problem involving both input-side activations and output-side gradients.
\item We introduce \mymethod{}, which uses orthogonal Procrustes alignment to transfer task vectors from a small calibration set using a single forward--backward pass, without any parameter updates.
\item Through extensive experiments in both computer vision and natural language processing, we demonstrate that \mymethod{} significantly outperforms existing methods and substantially narrows the gap to directly fine-tuning the target model.
\end{itemize}

\section{Related work}
\label{sec:related_work}
\paragraph{Model merging.}
Model merging aims to combine multiple task-specific models without additional training~\citep{ilharco2023editing,yadav2023ties,dare,localizing,tasksingular}. 
A representative approach is Task Arithmetic~\citep{ilharco2023editing}, which represents fine-tuned expertise as task vectors in parameter space and manipulates them through linear operations such as addition and negation. 
Subsequent methods, including TIES-Merging~\citep{yadav2023ties} and DARE~\citep{dare}, further improve merging by reducing interference among task vectors. 
Recent work further improves this paradigm by identifying consensus directions among task vectors~\citep{localizing} or by exploiting their low-rank singular structure to reduce task interference~\citep{tasksingular}.
These studies demonstrate that task vectors provide a useful abstraction for reusing and composing fine-tuned knowledge. 
However, most model-merging methods operate under a shared-parameter-space assumption, typically requiring task vectors to be defined relative to the same pre-trained model.
This limits their applicability when models differ in architecture, width, or initialization, and points to a broader question of how to relate models whose parameters are not directly aligned.

\paragraph{Weight and representation alignment.}
From the parameter-space perspective, Re-basin methods~\citep{gitrebasin, update} are based on the observation that models with the same architecture but different initializations can learn similar functions while representing them with different hidden-unit orderings. They address this parameter-space mismatch by exploiting permutation symmetries to map models into a shared basin.
However, these techniques are generally restricted to models with identical architectures and widths. 
Representation-level studies provide another perspective by comparing models through functional similarity rather than parameter correspondence~\citep{CKA,similarity,stitching,stitchingnew}.
Centered kernel alignment (CKA)~\citep{CKA} reveals shared internal structure across independently trained networks, while model stitching~\citep{similarity,stitching, stitchingnew} shows that such representations can sometimes be connected through simple mappings.
These studies suggest that independently trained models can exhibit compatible internal representations, motivating the use of representation alignment to translate source task vectors into target-compatible coordinates.

\paragraph{Task-specific parameter update transfer.} 
Recent work has explored transferring task-specific parameter updates, often referred to as task vectors~\citep{ilharco2023editing}, from a source model to a target pre-trained model without repeating full fine-tuning.
TransFusion~\citep{update} approaches this problem from a parameter-alignment perspective: it estimates permutation mappings between the source and target pre-trained models and applies the resulting re-basing transformation to the source task-vector. 
GradFix~\cite{gradfix} also operates on the task vector in parameter space, using one-step target-model gradients to construct a sign-consistency mask that suppresses update coordinates conflicting with the target loss landscape.
However, these methods rely on parameter-level compatibility, either through matched weights for permutation alignment or compatible tensor coordinates for coordinate-wise
masking.
To move beyond this parameter-wise compatibility requirement, THESEUS~\citep{theseus} characterizes a task update by the functional effect it induces on intermediate representations, using layer input and output activations to align update responses across models. 
Although this activation-based formulation flexibly establishes cross-model correspondences, its remaining gap to direct fine-tuning suggests that activation-level alignment alone may not fully capture the geometry of task-vector transfer.
By contrast, \mymethod{} revisits task-vector formation.
We show that a task vector can be derived as the accumulation of bilinear interactions between input-side activations and output-side gradients.
This perspective leads to a different transfer principle: rather than re-basing parameter coordinates, masking parameter entries, or matching activation responses alone, BiCo
transfers task vectors by aligning the two coordinate systems that define the bilinear update itself.


\label{sec:background}
\section{Preliminaries}
\subsection{Setup and notation}
Consider a pre-trained model with parameters $\vtheta_{\mathrm{pre}}$. When this model is adapted to a downstream task via fine-tuning, the resulting parameters are denoted as $\vtheta_{\mathrm{ft}}$. Following \citet{ilharco2023editing}, we define the task-vector $\boldsymbol{\tau}$ as the difference between the expert model and the pre-trained model:
\begin{equation}
    \boldsymbol{\tau} = \vtheta_{\mathrm{ft}} - \vtheta_{\mathrm{pre}}.
\end{equation}

In this work, we consider the problem of transferring a task-vector, available for a source pre-trained model $A$, to a different target pre-trained model $B$, with the goal of reusing the same task-specific knowledge without re-fine-tuning the target model.
We assume both models share the same architectural backbone (e.g., Transformer) but may differ in hidden width. 
For a specific linear layer or attention projection $l$, the source task-vector is defined as:
\begin{equation}
    \vtau_A^{(l)} = \vtheta_{A,\mathrm{ft}}^{(l)} - \vtheta_{A,\mathrm{pre}}^{(l)},
\end{equation}
where $\vtau_A^{(l)}
\in \mathbb{R}^{d_{\mathrm{out},A} \times d_{\mathrm{in},A}}$ while \(d_{\mathrm{in},A}\) and \(d_{\mathrm{out},A}\) denote the input and output dimensions of layer \(l\) in the source model \(A\). 
Our objective is to transfer the task-specific knowledge from $A$ to $B$ by constructing a target task-vector $\hat{\vtau}_B^{(l)} \in \mathbb{R}^{d_{\mathrm{out},B} \times d_{\mathrm{in},B}}$, where \(d_{\mathrm{in},B}\) and \(d_{\mathrm{out},B}\) denote the input and output dimensions of layer \(l\) in the target model \(B\). 
The goal is to ensure that the updated target layer, $\vtheta_{B, \mathrm{pre}}^{(l)} + \hat{\vtau}_B^{(l)}$, induces a functional effect analogous to the source expert layer, without requiring additional fine-tuning on model $B$.

\subsection{Sequence-length matching}
To handle activations from different models, we adopt the pre-processing steps described in \citet{theseus}. Given $N$ calibration inputs, we define the input activation of layer $l$ as:
\begin{equation}
    \mX_{A}^{(l)} \in \mathbb{R}^{N \times L_A \times d_{\mathrm{in},A}}, \quad \mX_{B}^{(l)} \in \mathbb{R}^{N \times L_B \times d_{\mathrm{in},B}} ,
    \label{eq:act_def}
\end{equation}
where $L_A$ and $L_B$ represent the respective sequence lengths. If $L_A \neq L_B$, we employ bilinear interpolation to align the sequence length of the source model to match $L_B$. By aggregating the batch and sequence dimensions into a total of $M = N \times L_B$ tokens, we obtain the flattened activation matrices $\bar\mX_{A}^{(l)} \in \mathbb{R}^{M \times d_{\mathrm{in},A}}, \quad \bar\mX_{B}^{(l)} \in \mathbb{R}^{M \times d_{\mathrm{in},B}}$.

\section{Method}
\label{sec:method}
\subsection{Revisiting task-vector formation}
\label{subsec:origin_task_vectors}

To transfer a task-vector $\vtau$ across different models, we revisit the fine-tuning update process to understand the formation of the task-vector at each layer. 
We consider a simplified case of the linear layer $\vy_t = \vx_t \vtheta_t^\top$ updated via standard gradient descent,
 where $\vtheta_t \in \mathbb{R}^{d_{\mathrm{out}} \times d_{\mathrm{in}}}$ is the weight matrix and $\vx_t\in \mathbb{R}^{1 \times d_{\mathrm{in}}}$ is the input activation at optimization step $t$.

The gradient of the loss $\mathcal{L}_t$ with respect to $\vtheta_t$ can be decomposed by the chain rule between the output-side gradient $\vg_t = \frac{\partial \mathcal{L}_t}{\partial \vy_t} \in \mathbb{R}^{1 \times d_{\mathrm{out}}}$ and the input activation $\vx_t$:
\begin{equation}
    \nabla_{\vtheta_t} \mathcal{L}_t
    =
    \frac{\partial \mathcal{L}_t}{\partial \vy_t}
    \left(
    \frac{\partial \vy_t}{\partial \vtheta_t}
    \right)
    =
    (\vg_t)^\top \vx_t .
    \label{eq:chain_rule}
\end{equation}
 
By accumulating these updates over $T$ optimization steps with a learning rate $\eta_t$, the task-vector $\vtau^{(l)} \in \mathbb{R}^{d_\mathrm{out} \times d_{\mathrm{in}}}$ for the layer $l$ is formulated as:
\begin{equation}
    \vtau^{(l)} 
    = \vtheta_{\mathrm{ft}}^{(l)} - \vtheta_{\mathrm{pre}}^{(l)}
    = \sum_{t=1}^{T} \Delta \vtheta_t^{(l)}
    =
    -\sum_{t=1}^{T} \eta_t \nabla_{\vtheta_t^{(l)}} \mathcal{L}_t
    =
    -\sum_{t=1}^{T} \eta_t ({\vg_t^{(l)}})^\top \vx_t^{(l)} .
    \label{eq:task_vector_decomp}
\end{equation}

Equation~\ref{eq:task_vector_decomp} suggests that a task-vector can be considered as a bilinear interaction coupling two factors: the input-side activation $\vx_t^{(l)}$  and the output-side gradient $\vg_t^{(l)}$. 
This motivates us to interpret cross-model task-vector transfer as reducing bilinear coordinate mismatch.



\subsection{Input-side coordinate alignment}
\label{subsec:inputside}
We now align the two coordinate spaces underlying the bilinear structure of task-vector formation. 
We first focus on the input side, where alignment should ideally reflect the activation factors accumulated along the fine-tuning trajectory. 
Prior work has shown that layer inputs remain temporally consistent during fine-tuning~\citep{WUDI}, suggesting that pre-trained activations may serve as a proxy for the trajectory activation factors. 
{
Following the input-consistency diagnostic of prior work~\citep{WUDI}, we further verify in Appendix~\ref{appendix:activation} that pre-trained activations $\vx_{\mathrm{pre}}$ remain close to the trajectory activations $\vx_t$ in our setting. 
This observation supports using $\vx_{\mathrm{pre}}$ as an efficient, non-iterative proxy for the input-side 
factors, avoiding the need to trace activations along the full fine-tuning trajectory.
}

Having established pre-trained activations as proxies for the input-side factors, we next ask how to relate the activation spaces of the source and target pre-trained models.
Prior work suggests that neural networks can learn comparable internal representations while expressing them in different coordinate systems~\citep{CKA,stitching,similarity}.
Modern pre-trained models are often trained on comparable large-scale corpora, suggesting that their feature spaces may share semantic structure~\citep{gadre2023datacomp,schuhmann2022laion5b}.
To examine this hypothesis, we compare the input activations of layer $l$ at the pre-trained initialization, denoted as
\(\bar\mX_{A, \mathrm{pre}}^{(l)} \in \mathbb{R}^{M \times d_{\mathrm{in},A}}\) and \(\bar\mX_{B, \mathrm{pre}}^{(l)} \in \mathbb{R}^{M \times d_{\mathrm{in},B}}\), using \(M\) calibration tokens obtained from the same \(N\) calibration inputs.
We evaluate representation similarity using centered kernel alignment (CKA)~\citep{CKA} and cosine similarity, which capture complementary notions of alignment.
CKA compares the sample-wise relational geometry of two representation spaces, so high CKA indicates that the two models organize samples similarly, even if their representations are expressed in different coordinate bases.
Cosine similarity, in contrast, compares corresponding activation vectors in their raw coordinates through a normalized dot product, making it sensitive to whether the corresponding hidden dimensions are aligned.
Since raw cosine similarity requires matching feature dimensions, we use same-width model pairs to assess coordinate-wise alignment.

\begin{figure}[t]
  \centering
  \includegraphics[width=0.80\linewidth]{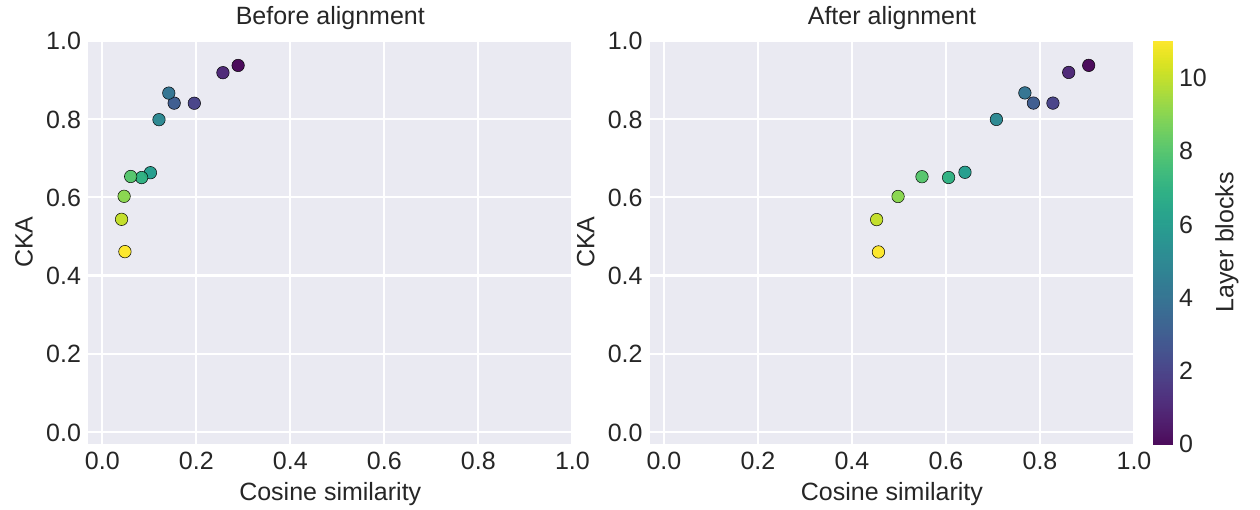}
  \vspace{-0.5em}
  \caption{\textbf{Representational similarity before and after input-side alignment.}
  We compare CKA and cosine similarity between pre-trained activations of same-width ViT-B/16 models with different pre-training distributions (A: Datacomp-XL $\rightarrow$ B: LAION-2B).
  (Left) before alignment. (Right) after Procrustes alignment.
  Results are averaged across 8 vision tasks and sub-layers within each residual block, using $N=100$ calibration samples to estimate the Procrustes map.}
  \label{fig:motivation}
\end{figure}

As shown in Figure~\ref{fig:motivation}, pre-trained models with different pre-training distributions exhibit high CKA across corresponding layers, indicating similar sample-wise relational geometry.
However, their activation cosine similarity remains extremely low, showing that the corresponding feature directions are not coordinate-wise aligned.
Moreover, high CKA also persists across different-width model pairs, as shown in Appendix~\ref{app:repr_similarity}.
This discrepancy suggests that pre-trained models can preserve similar representational geometry while expressing it in distinct coordinate systems.

Accordingly, we seek a transformation that aligns coordinate bases while preserving the shared representational geometry.
To this end, we adopt orthogonal Procrustes alignment~\citep{schonemann1966generalized}, which estimates the best orthogonality-constrained map between paired source and target representations. 
This allows us to align coordinates without distorting source-side norms, angles, or relational structure.
Given paired layer input activations $\bar\mX_{A,\mathrm{pre}}^{(l)}$ and $\bar\mX_{B,\mathrm{pre}}^{(l)}$, we seek an alignment matrix $\mR_{\mathrm{in}}^{(l)} \in \mathbb{R}^{d_{\mathrm{in}, A} \times d_{\mathrm{in}, B}}$ that maps the coordinates of $A$ to $B$.
For the width-expansion case where $d_{\mathrm{in},A} \le d_{\mathrm{in},B}$, we formulate this alignment as:
\begin{equation}
\mR_{\mathrm{in}}^{(l)} = \arg\min_{\mR} \|\bar\mX_{A, \mathrm{pre}}^{(l)} \mR - \bar\mX_{B, \mathrm{pre}}^{(l)}\|_F^2, \quad \text{s.t. } \mR \mR^\top = \mI_{d_{\mathrm{in},A}} .
\label{eq:procrustes_x}
\end{equation}
This problem admits a closed-form solution via Singular Value Decomposition (SVD). 
Specifically, letting $(\bar\mX_{A,\mathrm{pre}}^{(l)})^\top \bar\mX_{B,\mathrm{pre}}^{(l)} = U_{\mathrm{in}}\Sigma_{\mathrm{in}} V_{\mathrm{in}}^\top$, the optimal alignment matrix is given by $\mR_{\mathrm{in}}^{(l)} = U_{\mathrm{in}}V_{\mathrm{in}}^\top$. 
As shown in Figure~\ref{fig:motivation}, applying \(\mR_{\mathrm{in}}^{(l)}\) substantially increases activation cosine similarity while maintaining high CKA.
This suggests that the learned map improves coordinate-wise alignment while preserving the shared relational geometry.

\subsection{Output-side coordinate alignment}
\label{subsec:outputside}
The second factor in the bilinear update is the gradient with respect to the layer output, $g^{(l)} = \partial \mathcal{L} / \partial \vy^{(l)}$.
Unlike input activations, which characterize the representation geometry entering a layer, output-side gradients characterize the task-induced sensitivity of the loss in the layer output coordinates.
However, the input-only update remains in the source output coordinate system.
As a result, if only the input-side map is applied, the transferred update becomes $\hat{\vtau}^{(l)}_{B,\mathrm{in}} = \vtau^{(l)}_A \mR^{(l)}_{\mathrm{in}} \in\mathbb{R}^{d_{\mathrm{out},A}\times d_{\mathrm{in},B}}$, whereas the target weight matrix has shape $\vtheta_{B,\mathrm{pre}}^{(l)} \in \mathbb{R}^{d_{\mathrm{out},B}\times d_{\mathrm{in},B}}$.
Thus, when $d_{\mathrm{out},A}\neq d_{\mathrm{out},B}$, the input-only update cannot be directly added to the target weight matrix.
Even when the output dimensions match, dimensional compatibility alone does not guarantee coordinate compatibility:
the update can still be coordinate-wise misaligned because the two models may encode task-sensitive output directions under different bases.

In principle, a trajectory-level alignment of the output factor would require access to the gradients $\{\vg^{(l)}_{A,t}\}_{t=1}^T$ and $\{\vg^{(l)}_{B,t}\}_{t=1}^T$ accumulated throughout fine-tuning.
However, the source fine-tuning trajectory is no longer available after being summarized into the final task-vector $\vtau_A$,
while obtaining the corresponding target-side trajectory would require fine-tuning the target model.
Therefore, instead of matching trajectory gradients, we estimate the output-side coordinate correspondence from a single local measurement at the pre-trained initialization.
These gradients provide a local, task-aware signal of which output directions are sensitive to the downstream objective before any parameter update is applied.

Following the input-side alignment, we compute the gradients with respect to the layer outputs at the pre-trained initialization using the same set of $N$ calibration inputs.
To ensure consistency with the input-side processing, these gradients are reorganized in an identical manner into matrices $\bar\mG_{A, \mathrm{pre}}^{(l)} \in \mathbb{R}^{M \times d_{\mathrm{out},A}}$ and $\bar\mG_{B, \mathrm{pre}}^{(l)} \in \mathbb{R}^{M \times d_{\mathrm{out},B}}$.
Under the same width-expansion convention, the alignment is then formulated as an orthogonal Procrustes problem:
\begin{equation}
\mR_\mathrm{out}^{(l)} = \arg\min_{R} \|\bar\mG_{A,\mathrm{pre}}^{(l)} \mR - \bar\mG_{B,\mathrm{pre}}^{(l)}\|_F^2, \quad \text{s.t. } \mR \mR^\top = \mI_{d_{\mathrm{out},A}},
\label{eq:procrustes_g}
\end{equation}
Similar to the input-side alignment, this problem admits a closed-form solution via SVD.
Specifically, letting 
$(\bar\mG_{A,\mathrm{pre}}^{(l)})^\top 
\bar\mG_{B,\mathrm{pre}}^{(l)}
= U_{\mathrm{out}}\Sigma_{\mathrm{out}}V_{\mathrm{out}}^\top$,
the optimal output-side alignment matrix is given by
$\mR_{\mathrm{out}}^{(l)} = U_{\mathrm{out}}V_{\mathrm{out}}^\top$.

\subsection{Task-vector transfer via bilinear alignment}
\label{subsec:dualspace}
Before applying bilinear transfer, we note that the accumulated fine-tuning update is already contained in the source task-vector \(\vtau_A^{(l)}\). 
The single-pass calibration signals are therefore not used to construct \(\vtau_A^{(l)}\), but only to estimate the coordinate maps \(\mR_{\mathrm{in}}^{(l)}\) and \(\mR_{\mathrm{out}}^{(l)}\). 
With both alignment matrices determined, we align the source task-vector on both sides of its bilinear update structure.
The resulting task-vector $\hat{\vtau}_B^{(l)}\in
    \mathbb{R}^{d_{\mathrm{out},B}\times d_{\mathrm{in},B}}$  is obtained as
\begin{equation}
\begin{aligned}
    \hat{\vtau}_B^{(l)} 
    &:=
    -\sum_{t=1}^{T} \eta_t
    \left( \vg_{A,t}^{(l)} \mR_\mathrm{out}^{(l)} \right)^\top 
    \left( \vx_{A,t}^{(l)} \mR_{\mathrm{in}}^{(l)} \right) \\
    &=
    (\mR_\mathrm{out}^{(l)})^\top
    \left(
    -\sum_{t=1}^{T} \eta_t
    (\vg_{A,t}^{(l)})^\top \vx_{A,t}^{(l)}
    \right)
    \mR_{\mathrm{in}}^{(l)} \\
    &=
    (\mR_\mathrm{out}^{(l)})^\top \vtau_A^{(l)} \mR_{\mathrm{in}}^{(l)} .
\end{aligned}
\label{eq:full_transfer}
\end{equation}
A detailed description of the complete procedure is provided in Appendix~\ref{appendix:code}.
Finally, the total updated target model $\hat\vtheta_{B, \mathrm{ft}}$ is constructed by adding the aligned task vectors to its original pre-trained weights:
\begin{equation}
    \hat\vtheta_{B,\mathrm{ft}} = \vtheta_{B, \mathrm{pre}} + \hat{\vtau}_B .
    \label{eq:final_update}
\end{equation}
Notably, this update is performed without any additional hyperparameter tuning, such as searching for an optimal scaling factor, providing a robust and plug-and-play advantage for practical model transfer.

\paragraph{Effect of output-side alignment.}
\begin{wrapfigure}{r}{0.50\textwidth}
  \centering
  \vspace{-1.5em}
  \includegraphics[width=\linewidth]{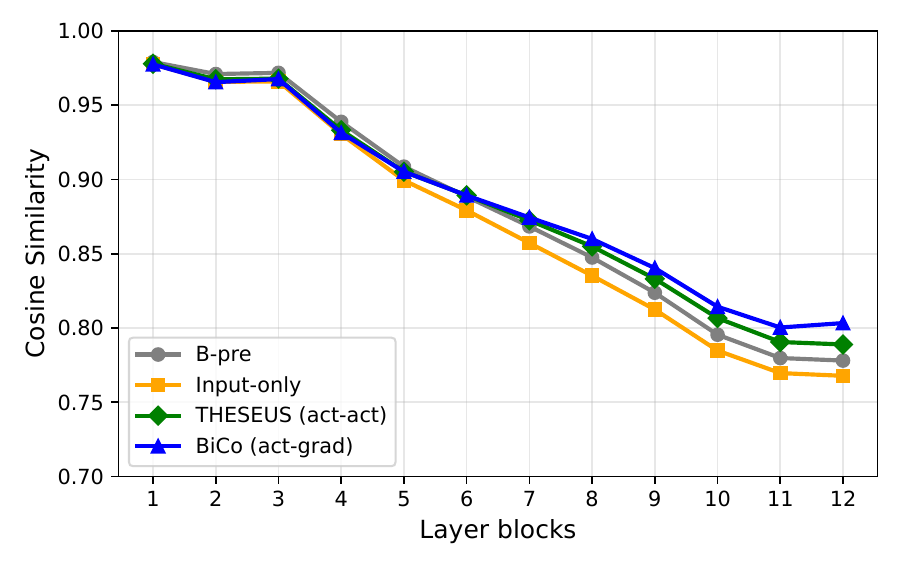}
  \vspace{-2em}
  \caption{
  \textbf{Layer-wise output similarity to the fine-tuned target.} 
    For ViT-B/16 (\textit{A}: Datacomp-XL → \textit{B}: LAION-2B), we compare layer-wise output activations of each strategy  with those of \(\vtheta_{B,\mathrm{ft}}\).
    \mymethod{} is closest, especially in deeper blocks.
    }
    \vspace{-1.0em}
  \label{fig:output_alignment}
\end{wrapfigure}

We analyze whether each transfer strategy induces layer-wise behavior similar to that of the oracle target model $\vtheta_{B,\mathrm{ft}}$, obtained by directly fine-tuning model $B$ on the task.
Since feature-level cosine similarity requires matched dimensionality, we conduct this analysis in the same-architecture setting using ViT-B/16 models.
To this end, we compare the output activations of transferred models with those of $\vtheta_{B,\mathrm{ft}}$.
As shown in Figure~\ref{fig:output_alignment}, the pre-trained target model (\textit{B}-pre) already has high cosine similarity to $\vtheta_{B,\mathrm{ft}}$, indicating that fine-tuning largely refines the existing output geometry.
However, input-only transfer $(\vtheta_{B,\mathrm{pre}}+\hat\vtau_{B,\mathrm{in}})$ degrades this similarity in deeper layers,  even falling below the pre-trained target model, suggesting that aligning only the input side is insufficient.
THESEUS improves over input-only transfer by matching input and output activations, but its layer outputs remain less aligned with $\vtheta_{B,\mathrm{ft}}$ than those of \mymethod{}, especially in later blocks.
This suggests that activation-based functional matching does not necessarily recover the output-side behavior induced by target fine-tuning.
By aligning output-side gradients, \mymethod{} produces layer responses that more closely follow the fine-tuned target model, supporting the effectiveness of gradient-based output-side alignment.

\section{Experiments}
We evaluate \mymethod{} across a broad suite of vision and natural language processing (NLP) benchmarks. 
Specifically, we conduct a comprehensive evaluation on the widely-used 8-Vision benchmark~\citep{ilharco2023editing} and five representative NLP tasks from the GLUE suite~\citep{GLUE}. Adopting the general evaluation framework established by THESEUS~\citep{theseus}, our experiments cover three representative transfer scenarios: (i) models with different widths, (ii) models with identical architectures, and (iii) models that differ in both width and depth. For each task, we estimate the alignment spaces from a small calibration set of \(N\) samples and directly apply the transferred task-vector without target-model training or task-specific hyperparameter tuning. Full experimental details are provided in Appendix~\ref{app:exp}.
\paragraph{Baselines.}
We report $\theta_{B}\, \textit{zero-shot}$ as the zero-shot target reference and $\theta_{B}\,\textit{fine-tune}$ as an oracle reference obtained by directly fine-tuning the target model.
We compare \mymethod{} with na\"ive task-vector transfer, which directly adds the source task-vector to the target model when shapes match and uses zero-padding otherwise, and with THESEUS~\citep{theseus} as the main training-free transfer baseline.
For identical-architecture transfer, we additionally include TransFusion~\citep{update} and GradFix~\citep{gradfix}, since their parameter-space transfer mechanisms require compatible model parameters.

\begin{table*}[t]
\caption{
\textbf{Width scaling (narrow → wide).}
Task-vector transfer from a ViT-B/16 model \textit{A} (LAION-2B) to a wider ViT-B/16-plus model \textit{B} (LAION-400M). \(N\) denotes the number of calibration samples used to estimate the Procrustes maps. $\Delta$Acc denotes accuracy gain over model \textit{B} zero-shot baseline.
}
\label{tab:table1}

\small
\renewcommand{\arraystretch}{1.05}
\setlength{\tabcolsep}{1.8pt}

\begin{tabular}{lcccccccccc}
\toprule
\textbf{Model} & $N$ & \textbf{EuroSAT} & \textbf{GTSRB} & \textbf{SVHN} & \textbf{RESISC45} & \textbf{DTD} & \textbf{Cars} & \textbf{MNIST} & \textbf{SUN397} & \textbf{AVG ($\Delta$Acc)} \\
\midrule
$\theta_B$ \textit{zero-shot} & -- & 42.51 & 49.61 & 37.77 & 65.95 & 55.42 & 84.50 & 57.02 & 68.50 & 57.66 {\color{black}(+0.00)} \\
$\theta_B$ \textit{fine-tune} & -- & 99.00 & 99.19 & 97.80 & 96.46 & 82.55 & 90.11 & 99.81 & 78.29 & 92.90 {\color{green!70!black}(+35.24)} \\
$\theta_A$ \textit{fine-tune} & -- & 94.67 & 99.03 & 97.89 & 96.81 & 83.46 & 92.10 & 99.72 & 79.35 & 92.88 {\color{green!70!black}(+35.22)} \\
$\theta_B + \tau_A^{\text{padded}}$ & -- & 46.41 & 38.21 & 27.27 & 62.57 & 53.56 & 82.00 & 51.70 & 68.29 & 53.75 {\color{red!80!black}(-3.91)} \\
\midrule
THESEUS & 10 & 44.03 & 51.48 & 45.53 & 66.07 & 55.53 & 84.31 & 58.62 & 68.67 & 59.28 {\color{green!70!black}(+1.62)} \\

\rowcolor[HTML]{FFF2CC}
\textbf{\mymethod} & 10 & 39.66 & 55.66 & 46.63 & 65.63 & 57.07 & 84.24 & 76.39 & 68.54 & \textbf{61.72} \textbf{{\color{green!70!black}(+4.06)}} \\

\midrule
THESEUS & 20 & 47.48&52.17&48.24 &67.50& 55.69& 84.14&54.93&68.51&59.83 {\color{green!70!black}(+2.17)} \\

\rowcolor[HTML]{FFF2CC}
\textbf{\mymethod} & 20 &68.81&57.06&67.12 & 67.92&58.24 &83.59 & 84.92& 68.80&\textbf{69.55} \textbf{{\color{green!70!black}(+11.89)}} \\

\midrule
THESEUS & 50 & 47.55 & 53.57 & 49.42 & 68.39 & 55.74 & 84.51 & 58.67 & 68.68 & 60.80 {\color{green!70!black}(+3.14)} \\

\rowcolor[HTML]{FFF2CC}
\textbf{\mymethod} & 50 & 78.25 & 60.99 & 78.14 & 72.39 & 59.25 & 85.23 & 91.97 & 68.98 & \textbf{74.40} \textbf{{\color{green!70!black}(+16.74)}} \\

\midrule
THESEUS & 100 & 48.85 & 53.82 & 49.02 & 68.98 & 55.85 & 84.60 & 62.00 & 68.71 & 61.47 {\color{green!70!black}(+3.81)} \\

\rowcolor[HTML]{FFF2CC}
\textbf{\mymethod} & 100 & 80.85 & 66.11 & 78.63 & 77.49 & 61.86 & 85.16 & 92.93 & 69.31 & \textbf{76.54} \textbf{{\color{green!70!black}(+18.88)}} \\

\bottomrule
\end{tabular}
\vspace{-1.5em}
\end{table*}

\subsection{Transfer across different widths}

\paragraph{Vision.} 
We first evaluate task-vector transfer across models of differing widths in a vision setting. Specifically, we transfer the task-vector $\tau_A$ learned by a source model \textit{A}, a CLIP ViT-B/16 pre-trained on LAION-2B, to a wider variant target model \textit{B}, the ViT-B/16-plus from OpenCLIP~\citep{clip} pre-trained on LAION-400M.
The results are summarized in Table~\ref{tab:table1}. 
We first observe that a na\"ive transfer approach using zero-padding ($\theta_B + \tau_A^{\text{padded}}$) to match dimensions results in a significant performance drop.
This confirms that task-vectors are not directly transferable across disparate parameter spaces without proper alignment. 
In contrast, \mymethod{} demonstrates strong transferability even in extremely low-data regimes.
With only 10 calibration samples ($N=10$), our method already surpasses the zero-shot performance of the target model.
\mymethod{} consistently benefits from larger calibration sets, yielding substantial gains over the zero-shot baseline and progressively approaching the fully fine-tuned target model. 
The performance gap against THESEUS becomes clear from $N=10$ to $N=20$ and further widens as $N$ increases, demonstrating the scalability of our bilinear coordinate alignment.

\paragraph{NLP.}
\begin{wraptable}{r}{0.48\textwidth}
\centering
\vspace{-0.2em}
\caption{\textbf{Performance of linear probing for task-vector transfer.} Source: T5-3B, Target: T5-Large.}
\label{tab:t5_linear_probe}
\vspace{-0.5em}
\scriptsize
\renewcommand{\arraystretch}{1.05}
\setlength{\tabcolsep}{1.6pt}

\resizebox{\linewidth}{!}{\begin{tabular}{lccccccc}
\toprule
\textbf{Encoder} & $N$ & \textbf{MNLI} & \textbf{QNLI} & \textbf{RTE} & \textbf{QQP} & \textbf{SNLI} & \textbf{AVG ($\Delta$Acc)} \\
\midrule
$\theta_A$ \textit{fine-tune} & All & 91.22 & 95.32 & 84.83 & 91.05 & 91.29 & 90.74 (-) \\
$\theta_B$ \textit{fine-tune} & All & 85.28 & 93.59 & 77.25 & 89.57 & 88.51 & 86.84 (-)\\
\midrule
$\theta_A$ & 50 & 39.34 & 75.58 & 58.12 & 68.62 & 53.21 & 58.97 {\color{green!70!black}(+4.27)} \\
$\theta_B$ & 50 & 38.74 & 69.50 & 57.40 & 62.56 & 45.28 & 54.70 {\color{black}(+0.00)} \\
THESEUS & 50 & 46.12 & 72.65 & 66.42 & 63.18 & 54.61 &  60.59 {\color{green!70!black}(+5.89)} \\
\rowcolor[HTML]{FFF2CC}
\textbf{\mymethod} & 50 & 46.32 & 76.71 & 67.87 & 83.40 & 60.80 & \textbf{67.02} \textbf{{\color{green!70!black}(+12.32)}} \\
\midrule
$\theta_A$ & 100 & 60.13 & 78.46 & 59.93 & 68.12 & 58.32 & 64.99 {\color{green!70!black}(+2.69)} \\
$\theta_B$ & 100 & 58.48 & 75.43 & 61.37 & 62.44 & 53.78 & 62.30 {\color{black}(+0.00)} \\
THESEUS & 100 & 60.93 & 78.34 & 66.78 & 63.19 & 56.38 & 65.12 {\color{green!70!black}(+2.82)} \\
\rowcolor[HTML]{FFF2CC}
\textbf{\mymethod} & 100 & 61.94 & 79.33 & 67.50 & 84.31 & 58.43 & \textbf{70.30} \textbf{{\color{green!70!black}(+8.00)}} \\
\midrule
$\theta_A$ & 200 & 62.49 & 81.18 & 65.70 & 86.23& 65.94 & 72.31 {\color{green!70!black}(+1.21)} \\
$\theta_B$ & 200 & 60.83 & 82.45 & 63.54 & 85.52 & 63.18 & 71.10 {\color{black}(+0.00)} \\
THESEUS & 200 & 69.92 & 83.63 & 68.23 & 88.09 & 64.31 & 74.83 {\color{green!70!black}(+3.73)} \\
\rowcolor[HTML]{FFF2CC}
\textbf{\mymethod} & 200 & 71.06 & 85.74 & 68.95 & 88.25 & 66.59 & \textbf{76.11} \textbf{{\color{green!70!black}(+5.01)}} \\
\bottomrule
\end{tabular}}
\vspace{-2.0em}
\end{wraptable}
To assess generalization to language models, we evaluate \mymethod{} using T5 architectures~\citep{t5}, with T5-3B as the source model \textit{A}, and T5-Large as the target model \textit{B}.
Following the protocol of THESEUS~\citep{theseus}, we transfer the encoder task-vector \(\tau_A\) and evaluate the transferred encoder via linear probing~\citep{linearprob}.
We vary the number of calibration samples $N$ used to estimate the alignment maps and to train the probe, ensuring an identical data budget for all compared methods. 
As shown in Table~\ref{tab:t5_linear_probe}, \mymethod{} demonstrates robust improvements in linear probing accuracy across all tested benchmarks and data regimes. 
We observe that the overall performance scales up as the number of samples $N$ increases, highlighting that the effectiveness of \mymethod{} successfully extends beyond the computer vision domain to NLP tasks.

\vspace{-0.5em}
\begin{table}[t]
\centering
\caption{
\textbf{Identical architecture transfer.}
Task-vector transfer between ViT-B/16 models with identical architectures but different pretraining (\textit{A}: Datacomp-XL → \textit{B}: LAION-2B). \(K\) denotes the number of examples per class used to estimate the Procrustes maps. $\Delta$Acc denotes accuracy gain over model \textit{B} zero-shot baseline.
}
\label{tab:table3}
\small
\renewcommand{\arraystretch}{1.05}
\setlength{\tabcolsep}{1.8pt} 
\begin{tabular}{lccccccccccc}
\toprule
\textbf{Model} & $K$ &  \textbf{EuroSAT} & \textbf{GTSRB} & \textbf{SVHN} & \textbf{RESISC45} & \textbf{DTD} & \textbf{Cars} & \textbf{MNIST} & \textbf{SUN397} & \textbf{AVG ($\Delta$Acc)}\\
\midrule
$\theta_B$ \textit{zero-shot} & -- & 50.15 & 48.35 & 50.05 & 68.22 & 55.96 & 88.56 & 65.79 & 70.44 & 62.19 {\color{black}(+0.00)} \\
$\theta_B$ \textit{fine-tune} & -- & 94.67 & 99.03 & 97.89 & 96.81 & 83.46 & 92.10 & 99.72 & 79.35 & 92.88 {\color{green!70!black}(+30.69)} \\
$\theta_B + \tau_A$ & -- & 36.07 & 33.96 & 20.33 & 63.16 & 54.73 & 86.82 & 72.09 & 66.74 & 54.24 {\color{red!80!black}(-7.95)} \\
TransFusion & -- &  50.30 & 49.48 &52.57 & 68.27 & 56.65 & 88.60 & 73.51 & 70.51 & 63.74 {\color{green!70!black}(+1.55)} \\
\midrule
GradFix & 1 & 69.67 & 60.32 & 57.28 & 73.02 & 56.91 & 85.50 & 87.66 & 71.34 & 70.21 {\color{green!70!black}(+8.02)} \\
THESEUS & 1 & 50.18 & 60.23 & 60.77 & 70.63 & 56.80 & 87.45 & 84.04 & 70.80 & 67.61 {\color{green!70!black}(+5.42)} \\
\rowcolor[HTML]{FFF2CC}
\textbf{\mymethod} & 1 & 68.00 & 74.47 & 74.94 & 71.98 & 63.24 & 87.76 & 91.48 & 72.21 & \textbf{75.51} \textbf{{\color{green!70!black}(+13.32)}} \\
\midrule
GradFix & 2 & 64.15 & 62.95 & 63.42 & 73.18 & 57.18 & 85.50 & 88.46 & 71.64 & 70.81 {\color{green!70!black}(+8.62)} \\
THESEUS & 2 & 55.40 & 57.90 & 56.78 & 71.49 & 56.91 & 87.61 & 87.80 & 70.72 & 68.08 {\color{green!70!black}(+5.89)} \\
\rowcolor[HTML]{FFF2CC}
\textbf{\mymethod} & 2 & 73.96 & 79.50 & 71.32 & 78.60 & 67.07 & 87.53 & 91.39 & 72.44 & \textbf{77.73} \textbf{{\color{green!70!black}(+15.54)}} \\
\midrule
GradFix & 5 & 65.30 & 63.78 & 65.06 & 71.85 & 57.18 & 82.92 & 89.48 & 71.08 & 70.83 {\color{green!70!black}(+8.64)} \\
THESEUS & 5 & 56.55 & 58.47 & 60.66 & 71.93 & 58.03 & 87.10 & 87.47 & 70.81 & 68.88 {\color{green!70!black}(+6.69)} \\
\rowcolor[HTML]{FFF2CC}
\textbf{\mymethod} & 5 & 78.25 & 81.96 & 72.69 & 85.65 & 68.45 & 87.68 & 92.21 & 73.11 & \textbf{80.00} \textbf{{\color{green!70!black}(+17.81)}} \\
\bottomrule
\end{tabular}
\vspace{-1.5em}
\end{table}
\subsection{Transfer between identical architectures}
\label{identical}
We further evaluate \mymethod{} in an identical-architecture setting, where both source and target models use ViT-B/16 but are pre-trained on different data distributions.
Specifically, we transfer task vectors from a source model \textit{A} pre-trained on Datacomp-XL to a target model \textit{B} pre-trained on LAION-2B.
For methodological consistency,  we follow the evaluation protocol of GradFix~\citep{gradfix} and use a standard $K$-shot setting, where $K$ denotes the number of examples per class used for coordinate alignment.
As shown in Table~\ref{tab:table3}, na\"ively applying the source task-vector to the target model, $\theta_B + \tau_A$, degrades performance below the zero-shot baseline, confirming the need for coordinate alignment even when architectures match.
Among alignment methods, TransFusion, which relies on permutation-based symmetries, shows limited effectiveness.
Similarly, THESEUS and GradFix provide only modest improvements over the baseline, struggling to fully capture the task-specific expertise.
By contrast, \mymethod{} achieves stronger performance with fewer calibration samples and continues to improve as $K$ increases, showing that aligning both input activations and output-side gradients enables more effective task-vector transfer and steadily narrows the gap to direct target fine-tuning.

\begin{table}[t]
\caption{\textbf{Depth and width scaling.}
Task-vector transfer from a  ViT-B/16 model to ViT-L/14 model, both pretrained on Datacomp-XL. \(N\) denotes the number of calibration samples used to estimate the Procrustes maps. $\Delta$Acc denotes accuracy gain over model \textit{B} zero-shot baseline.}
\centering
\small
\renewcommand{\arraystretch}{1.05}
\setlength{\tabcolsep}{1.8pt} 
\begin{tabular}{lcccccccccc}
\toprule
\textbf{Model} & $N$ & \textbf{EuroSAT} &  \textbf{GTSRB} & \textbf{SVHN} & \textbf{RESISC45} & \textbf{DTD} & \textbf{Cars} & \textbf{MNIST} & \textbf{SUN397} & \textbf{AVG ($\Delta$Acc)} \\
\midrule
$\theta_B$ \textit{zero-shot} & -- &  64.11& 58.91&	67.67&	72.60&	66.96&	93.06&	86.63&	73.77& 70.29 {\color{black}(+0.00)}  \\
$\theta_B$ \textit{fine-tune} & -- & 99.33&	99.29&	98.21&	97.58&	86.11&	95.22&	99.74&	82.40 &94.74 {\color{green!70!black}(+24.45)}\\
$\theta_A$ \textit{fine-tune} & -- & 98.96&	99.06&	98.02&	96.71&	84.25&	92.50&	99.77&	79.25& 93.57 {\color{green!70!black}(+23.28)} \\
$\theta_B + \tau_A^{\text{padded}}$ & -- & 64.07&58.38& 68.14&72.87& 66.70& 93.11 &86.98& 73.78& 73.00 {\color{green!70!black}(+2.71)} \\
\midrule
THESEUS & 100 & 66.96&	62.55&	71.05&	74.33&	67.81&	93.25&	89.88&	73.77&74.95 {\color{green!70!black}(+4.66)}\\
\rowcolor[HTML]{FFF2CC}
\textbf{\mymethod} & 100 & 80.44&	78.63& 86.10 & 82.33 & 71.17 & 93.80 & 98.27 & 74.70 & \textbf{83.18} \textbf{{\color{green!70!black}(+12.89)}}\\
\toprule
\end{tabular} 
\vspace{-2em}
\label{tab:table4}
\end{table}

\subsection{Transfer across width and depth}
To evaluate the structural generalization of our approach, we consider a challenging setting where the source and target models differ in both width and depth: ViT-B/16 as the source model \textit{A} and ViT-L/14 as the target model \textit{B}, both pre-trained on Datacomp-XL.
To handle depth mismatch, we use a simple discrete layer-index matching strategy. 
Given a source model with $D_A$ layers and a target model with $D_B$ layers, each target layer $j$ is matched to a source layer $i(j)=\operatorname{round}\left(\frac{j(D_A-1)}{D_B-1}\right)$,
where $i(j) \in \{0,\ldots,D_A-1\}$.
This nearest-neighbor rule preserves layer ordering and naturally reuses source layers when transferring from a shallower to a deeper target.
Additional results for deeper-to-shallower transfer and different pre-training distributions are provided in Appendix~\ref{extend}.
For each matched pair $(i(j),j)$, we apply Procrustes alignment to transfer the source-layer task-vector into the corresponding target-layer coordinate space.
Table~\ref{tab:table4} shows that \mymethod{} remains effective under simultaneous depth and width mismatch. 
Despite using only a simple discrete layer-matching strategy, \mymethod{} improves over the target zero-shot baseline by more than 10$\%$ on average, demonstrating its structural generalizability across models with mismatched depths and widths.

\vspace{-0.5em}
\begin{wraptable}{r}{0.50\textwidth}
\vspace{-1.5em}
\centering
\caption{
\textbf{Comparison of alignment strategies.}
Average accuracy across 8 vision tasks for ViT-B/16 (\textit{A}: Datacomp-XL→ \textit{B}: LAION-2B) task-vector transfer.
$N$ denotes the number of calibration samples.
}
\label{tab:ablation}
\vspace{-0.5em}
\small
\renewcommand{\arraystretch}{1.05}
\setlength{\tabcolsep}{1.6pt}
\begin{tabular}{lccccc}
\toprule
\textbf{Alignment} & \textbf{$N=10$} & \textbf{$N=20$} & \textbf{$N=50$} & \textbf{$N=100$} \\
\midrule
\textit{Input-only}  & 61.04 & 61.10 & 61.19 & 61.18 \\
\textit{Output-only} & 64.75 & 67.69 & 68.84 & 70.45  \\
\textit{Gradient-only}  & 62.31 & 63.09 & 64.46 & 65.55 \\
THESEUS & 65.78 & 67.38 & 68.54& 69.03\\
\midrule
\rowcolor[HTML]{FFF2CC}
\textbf{\mymethod} & \textbf{67.54} & \textbf{73.06} & \textbf{76.40} & \textbf{79.66} \\
\bottomrule
\end{tabular}
\vspace{-1.0em}
\end{wraptable}
\subsection{Analysis of bilinear alignment}
To investigate the importance of each alignment in \mymethod{}, we conduct an ablation study in the identical-architecture setting, transferring task vectors between ViT-B/16 models (\textit{A}: Datacomp-XL $\rightarrow$ \textit{B}: LAION-2B).
Table~\ref{tab:ablation} reports the average accuracy across eight vision tasks for varying numbers of samples $N$.
We evaluate four variants: 
(i) \textit{Input-only}, which aligns only input-side activations; 
(ii) \textit{Output-only}, which aligns only output-side gradients; 
(iii) \textit{Gradient-only}, which utilizes gradients instead of activations for the input-side alignment; and
(iv) THESEUS, which uses layer input and output activations for both-side alignment.
Notably, the \textit{Input-only} variant yields performance that is even lower than the zero-shot performance of the pre-trained target model (62.19$\%$) in Table~\ref{tab:table3}. This result is consistent with Figure~\ref{fig:output_alignment}, which shows that \textit{Input-only} transfer degrades feature similarity in deep layers even below the initial pre-trained model.
Moreover, \textit{Output-only} performs comparably to THESEUS, indicating the importance of our output-side gradient alignment.
The weak performance of \textit{Gradient-only}, which replaces input-side activation alignment with gradient-based alignment, shows that gains of \mymethod{} do not come merely from using gradients, but from aligning activations and gradients according to their distinct bilinear roles.

\section{Conclusion}
\label{sec:conclusion}
We presented \mymethod{}, a training-free framework for transferring task vectors across pre-trained model variants. 
By interpreting task vectors as bilinear interactions between input-side activations and output-side gradients, \mymethod{} aligns both coordinate spaces with Procrustes mappings estimated from a small calibration set.
Across vision and NLP benchmarks, \mymethod{} consistently outperforms na\"ive transfer and existing training-free baselines under variations in width, depth, and pre-training configuration.
These results highlight the importance of preserving the bilinear structure of fine-tuning updates for scalable reuse of task-specific knowledge.

\newpage
\bibliographystyle{plainnat}
\bibliography{main}

@inproceedings{WUDI,
author={Runxi Cheng and Feng Xiong and Yongxian Wei and Wanyun Zhu and Chun Yuan},
title={Whoever Started the Interference Should End It: Guiding Data-Free Model Merging via Task Vectors},
booktitle={ICML},
year={2025}
}

@inproceedings{CKA,
author={Simon Kornblith and Mohammad Norouzi and Honglak Lee and Geoffrey Hinton},
title={Similarity of Neural Network Representations Revisited},
booktitle={ICML},
year={2019}
}

@inproceedings{stitching,
author={Yamini Bansal and Preetum Nakkiran and Boaz Barak},
title={Revisiting Model Stitching to Compare Neural Representations},
booktitle={NeurIPS},
year={2021}
}

@inproceedings{similarity,
author={Adrián Csiszárik and Péter Kőrösi-Szabó and Ákos K. Matszangosz and Gergely Papp and Dániel Varga},
title={Similarity and Matching of Neural Network Representations},
booktitle={NeurIPS},
year={2021}
}

@article{theseus,
  title={Transporting Task Vectors across Different Architectures without Training},
  author={Filippo Rinaldi and Aniello Panariello and Giacomo Salici and Angelo Porrello and Simone Calderara},
  journal={arXiv preprint arXiv:2602.12952},
  year={2026}
}

@inproceedings{ilharco2023editing,
author = {Ilharco, Gabriel and Tulio Ribeiro, Marco and Wortsman, Mitchell and Gururangan, Suchin and Schmidt, Ludwig and Hajishirzi, Hannaneh and Farhadi, Ali},
title = {Editing models with task arithmetic},
booktitle = {ICLR},
year = {2023}
}

@inproceedings{gitrebasin,
author = {Samuel K. Ainsworth and Jonathan Hayase and Siddhartha Srinivasa},
title = {Git Re-Basin: Merging Models modulo Permutation Symmetries},
booktitle = {ICLR},
year = {2023}
}

@inproceedings{update,
author = {Filippo Rinaldi and Giacomo Capitani and Lorenzo Bonicelli and Donato Crisostomi and Federico Bolelli and Elisa Ficarra and Emanuele Rodolà and Simone Calderara and Angelo Porrello},
title = {Update Your Transformer to the Latest Release: Re-Basin of Task Vectors},
booktitle = {ICML},
year = {2025}
}

@inproceedings{gradfix,
author = {Filippo Rinaldi and Aniello Panariello and Giacomo Salici and Fengyuan Liu and Marco Ciccone and Angelo Porrello and Simone Calderara},
title = {Gradient-Sign Masking for Task Vector Transport Across Pre-Trained Models},
booktitle = {ICLR},
year = {2026}
}

@inproceedings{transformer,
title={An Image is Worth 16x16 Words: Transformers for Image Recognition at Scale},
author={Alexey Dosovitskiy and Lucas Beyer and Alexander Kolesnikov and Dirk Weissenborn and Xiaohua Zhai and Thomas Unterthiner and Mostafa Dehghani and Matthias Minderer and Georg Heigold and Sylvain Gelly and Jakob Uszkoreit and Neil Houlsby},
booktitle={ICLR},
year={2021} 
}

@article{achiam2023gpt,
  title={GPT-4 technical report},
  author={Achiam, Josh and Adler, Steven and Agarwal, Sandhini and Ahmad, Lama and Akkaya, Ilge and Aleman, Florencia Leoni and Almeida, Diogo and Altenschmidt, Janko and Altman, Sam and Anadkat, Shyamal and others},
  journal={arXiv preprint arXiv:2303.08774},
  year={2023},
}

@article{touvron2023llama,
  title={LLaMA: Open and Efficient Foundation Language Models},
  author={Touvron, Hugo and Lavril, Thibaut and Izacard, Gautier and Martinet, Xavier and Lachaux, Marie-Anne and Lacroix, Timothée and Rozière, Baptiste and Goyal, Naman and Hambro, Eric and Azhar, Faisal and Rodriguez, Aurelien and Joulin, Armand and Grave, Edouard and Lample, Guillaume},
  journal={arXiv preprint arXiv:2302.13971},
  year={2023},
}

@inproceedings{wei2022finetuned,
  title={Finetuned Language Models Are Zero-Shot Learners},
  author={Wei, Jason and Bosma, Maarten and Zhao, Vincent Y. and Guu, Kelvin and Yu, Adams Wei and Lester, Brian and Du, Nan and Dai, Andrew M. and Le, Quoc V.},
  booktitle= {ICLR},
  year={2022}
}

@inproceedings{wortsman2022robust,
  title={Robust fine-tuning of zero-shot models},
  author={Wortsman, Mitchell and Ilharco, Gabriel and Kim, Jong Wook and Li, Mike and Kornblith, Simon and Roelofs, Rebecca and Gontijo-Lopes, Raphael and Hajishirzi, Hannaneh and Farhadi, Ali and Namkoong, Hongseok and Schmidt, Ludwig},
  booktitle={CVPR},
  year={2022}
}

@inproceedings{brown2020language,
  title={Language Models are Few-Shot Learners},
  author={Brown, Tom B. and Mann, Benjamin and Ryder, Nick and Subbiah, Melanie and Kaplan, Jared and Dhariwal, Prafulla and Neelakantan, Arvind and Shyam, Pranav and Sastry, Girish and Askell, Amanda and Agarwal, Sandhini and Herbert-Voss, Ariel and Krueger, Gretchen and Henighan, Tom and Child, Rewon and Ramesh, Aditya and Ziegler, Daniel M. and Wu, Jeffrey and Winter, Clemens and Hesse, Christopher and Chen, Mark and Sigler, Eric and Litwin, Mateusz and Gray, Scott and Chess, Benjamin and Clark, Jack and Berner, Christopher and McCandlish, Sam and Radford, Alec and Sutskever, Ilya and Amodei, Dario},
  booktitle={NeurIPS},
  year={2020}
}

@inproceedings{yadav2023ties,
title={TIES-Merging: Resolving Interference When Merging Models},
author={Yadav, Prateek and Tam, Derek and Choshen, Leshem and Raffel, Colin and Bansal, Mohit},
booktitle={NeurIPS},
year={2023}
}

@inproceedings{dare,
title={Language Models are Super Mario: Absorbing Abilities from Homologous Models as a Free Lunch},
author={Le Yu and Bowen Yu and Haiyang Yu and Fei Huang and Yongbin Li},
booktitle={ICML},
year={2024}
}

@inproceedings{stitchingnew,
title={Revisiting Model Stitching In the Foundation Model Era},
author={Zheda Mai and Ke Zhang and Fu-En Wang and Zixiao Ken Wang and Albert Y. C. Chen and Lu Xia and Min Sun and Wei-Lun Chao and Cheng-Hao Kuo},
booktitle={CVPR},
year={2026}
}

@inproceedings{clip,
title={Reproducible scaling laws for contrastive language-image learning},
author={Mehdi Cherti and Romain Beaumont and Ross Wightman and Mitchell Wortsman and Gabriel Ilharco and Cade Gordon and Christoph Schuhmann and Ludwig Schmidt and Jenia Jitsev},
booktitle={CVPR},
year={2023}
}

@inproceedings{GLUE,
title={GLUE: A Multi-Task Benchmark and Analysis Platform for Natural Language Understanding},
author={Alex Wang and Amanpreet Singh and Julian Michael and Felix Hill and Omer Levy and Samuel R. Bowman},
booktitle={ICLR},
year={2019}
}

@article{t5,
  title={Exploring the Limits of Transfer Learning with a Unified Text-to-Text Transformer},
  author={Colin Raffel and Noam Shazeer and Adam Roberts and Katherine Lee and Sharan Narang and Michael Matena and Yanqi Zhou and Wei Li and Peter J. Liu},
  journal={Journal of Machine Learning Research},
  year={2020}
}

@article{linearprob,
  title={Understanding Intermediate Layers Using Linear Classifier Probes},
  author={Guillaume Alain and Yoshua Bengio},
  journal={arXiv preprint arXiv:1610.01644},
  year={2017}
}

@inproceedings{gadre2023datacomp,
  title={DataComp: In Search of the Next Generation of Multimodal Datasets},
  author={Samir Yitzhak Gadre and Gabriel Ilharco and Alex Fang and Jonathan Hayase and Georgios Smyrnis and Thao Nguyen and Ryan Marten and Mitchell Wortsman and Dhruba Ghosh and Jieyu Zhang and Eyal Orgad and Rahim Entezari and Giannis Daras and Sarah Pratt and Vivek Ramanujan and Yonatan Bitton and Kalyani Marathe and Stephen Mussmann and Richard Vencu and Mehdi Cherti and Ranjay Krishna and Pang Wei Koh and Olga Saukh and Alexander Ratner and Shuran Song and Hannaneh Hajishirzi and Ali Farhadi and Romain Beaumont and Sewoong Oh and Alex Dimakis and Jenia Jitsev and Yair Carmon and Vaishaal Shankar and Ludwig Schmidt},
  booktitle={NeurIPS},
  year={2023}
}

@inproceedings{tasksingular,
  title={Task Singular Vectors: Reducing Task Interference in Model Merging},
  author={Antonio Andrea Gargiulo and Donato Crisostomi and Maria Sofia Bucarelli and Simone Scardapane and Fabrizio Silvestri and Emanuele Rodolà},
  booktitle={CVPR},
  year={2025}
}

@inproceedings{localizing,
  title={Localizing Task Information for Improved Model Merging and Compression},
  author={Ke Wang and Nikolaos Dimitriadis and Guillermo Ortiz-Jimenez and François Fleuret and Pascal Frossard},
  booktitle={ICML},
  year={2024}
}

@article{schonemann1966generalized,
  title={A generalized solution of the orthogonal Procrustes problem},
  author={Sch{\"o}nemann, Peter H.},
  journal={Psychometrika},
  year={1966}
}

@inproceedings{schuhmann2022laion5b,
  title={LAION-5B: An open large-scale dataset for training next generation image-text models},
  author={Schuhmann, Christoph and Beaumont, Romain and Vencu, Richard and Gordon, Cade and Wightman, Ross and Cherti, Mehdi and Coombes, Theo and Katta, Aarush and Mullis, Clayton and Wortsman, Mitchell and Schramowski, Patrick and Kundurthy, Srivatsa and Crowson, Katherine and Schmidt, Ludwig and Kaczmarczyk, Robert and Jitsev, Jenia},
  booktitle={NeurIPS},
  year={2022}
}

@article{eurosat,
  title={Eurosat: A novel dataset and deep learning benchmark for land use and land cover classification},
  author={Helber, Patrick and Bischke, Benjamin and Dengel, Andreas and Borth, Damian},
  journal={Journal of Selected Topics in Applied Earth Observations and Remote Sensing},
  year={2019},
}

@inproceedings{gtsrb,
  title={The German Traffic Sign Recognition Benchmark: A multi-class classification competition},
  author={Stallkamp, Johannes and Schlipsing, Marc and Salmen, Jan and Igel, Christian},
  booktitle={IJCNN},
  year={2011}
}

@inproceedings{svhn,
  title={Reading digits in natural images with unsupervised feature learning},
  author={Netzer, Yuval and Wang, Tao and Coates, Adam and Bissacco, Alessandro and Wu, Bo and Ng, Andrew Y.},
  booktitle={NIPS Workshop on Deep Learning and Unsupervised Feature Learning},
  year={2011}
}

@article{resics45,
 title = {Remote Sensing Image Scene Classification: Benchmark and State of the Art},
 author = {Gong Cheng and Junwei Han and Xiaoqiang Lu},
 journal = {Proceedings of the IEEE},
 year = {2017}
}

@inproceedings{dtd,
 title = {Describing Textures in the Wild},
 author = {Mircea Cimpoi and Subhransu Maji and Iasonas Kokkinos and S. Mohamed and A. Vedaldi},
 booktitle = {CVPR},
 year = {2014}
}

@inproceedings{cars,
  title={3D Object Representations for Fine-Grained Categorization}, 
  author={Krause, Jonathan and Stark, Michael and Deng, Jia and Fei-Fei, Li},
  booktitle={ICCV Workshop}, 
  year={2013},
}

@article{mnist,
  title={Gradient-based learning applied to document recognition}, 
  author={Lecun, Y. and Bottou, L. and Bengio, Y. and Haffner, P.},
  journal={Proceedings of the IEEE}, 
  year={1998}
}

@inproceedings{sun397,
 title={SUN database: Large-scale scene recognition from abbey to zoo},
 author={Xiao, Jianxiong and Hays, James and Ehinger, Krista A. and Oliva, Aude and Torralba, Antonio},
 booktitle={CVPR},
 year={2010}
}

@inproceedings{snli,
  title={A large annotated corpus for learning natural language inference},
  author={Samuel R. Bowman and Gabor Angeli and Christopher Potts and Christopher D. Manning},
  booktitle={EMNLP},
  year={2015}
}

@inproceedings{mnli,
    title = {A Broad-Coverage Challenge Corpus for Sentence Understanding through Inference},
    author = {Williams, Adina  and Nangia, Nikita  and Bowman, Samuel},
    booktitle = {NAACL-HLT},
    year={2018}
}

@article{qqp,
  title={Natural language understanding with the quora question pairs dataset},
  author={Lakshay Sharma and Laura Graesser and Nikita Nangia and Utku Evci},
  journal={arXiv preprint arXiv:1907.01041},
  year={2019}
}

@article{cost,
    title = {Scaling Laws for Neural Language Models},
    author = {Jared Kaplan and Sam McCandlish and Tom Henighan and Tom B. Brown and Benjamin Chess and Rewon Child and Scott Gray and Alec Radford and Jeffrey Wu and Dario Amodei
    },
    journal={arXiv preprint arXiv:2001.08361},
  year={2020}
}

\newpage

\setcounter{section}{0}
\setcounter{subsection}{0}
\setcounter{figure}{0}
\setcounter{table}{0}
\setcounter{equation}{0}

\renewcommand{\thesection}{\Alph{section}}
\renewcommand{\thesubsection}{\thesection.\arabic{subsection}}

\renewcommand{\thefigure}{\Alph{figure}}
\renewcommand{\thetable}{\Alph{table}}
\renewcommand{\theequation}{\Alph{equation}}

\renewcommand{\theHsection}{app.\Alph{section}}
\renewcommand{\theHsubsection}{app.\Alph{section}.\arabic{subsection}}
\renewcommand{\theHfigure}{app.\Alph{figure}}
\renewcommand{\theHtable}{app.\Alph{table}}
\renewcommand{\theHequation}{app.\Alph{equation}}

\section*{Table of contents}
We provide the following items in this Appendix:
\begin{itemize}
    \item (\ref{appendix:code}) Pseudocode of \mymethod{}
    \item (\ref{appendix:activation}) Activation consistency along fine-tuning trajectories
    \item (\ref{app:repr_similarity}) Cross-width representational similarity analysis
    \item (\ref{app:orthogonal_preservation}) Geometry preservation of orthogonal alignment
    \item (\ref{app:exp}) Experimental details
    \item (\ref{app:addexps}) Additional experiments
    \item (\ref{app:computation}) Computational cost analysis
    \item (\ref{app:limitations}) Limitations
    \item (\ref{app:borader}) Broader impacts
\end{itemize}
\section{Pseudocode of BiCo}
\label{appendix:code}
\begin{algorithm}[ht]\small
    \caption{BiCo: Training-Free Task-Vector Transfer}\label{alg}
    \begin{algorithmic}[1]
    \REQUIRE Source model $\theta_{A,\mathrm{pre}}, \theta_{A,\mathrm{ft}}$, target model $\theta_{B,\mathrm{pre}}$, calibration dataset $\mathcal{D}$
    \FOR{each layer $l \in \{1, \ldots, L\}$}
        \STATE Compute source task-vector: $\tau_A^{(l)} \leftarrow \theta_{A,\mathrm{ft}}^{(l)} - \theta_{A,\mathrm{pre}}^{(l)}$
        \STATE Obtain input activations $X_{A,\mathrm{pre}}^{(l)}, X_{B,\mathrm{pre}}^{(l)}$ from $\mathcal{D}$
        \STATE Obtain output-side gradients $G_{A,\mathrm{pre}}^{(l)}, G_{B,\mathrm{pre}}^{(l)}$ from $\mathcal{D}$
        \STATE Interpolate sequence lengths if needed and flatten batch/token dimensions.
        \STATE Compute cross-covariances: \\
        \quad $C_{\mathrm{in}} \leftarrow (X_{A,\mathrm{pre}}^{(l)})^\top X_{B,\mathrm{pre}}^{(l)}, \quad C_{\mathrm{out}} \leftarrow (G_{A,\mathrm{pre}}^{(l)})^\top G_{B,\mathrm{pre}}^{(l)}$
        \STATE Compute SVD: \\
        \quad $C_{\mathrm{in}} = U_{\mathrm{in}} \Sigma_{\mathrm{in}} V_{\mathrm{in}}^\top, \quad C_{\mathrm{out}} = U_{\mathrm{out}} \Sigma_{\mathrm{out}} V_{\mathrm{out}}^\top$
        \STATE Solve orthogonal Procrustes problems: \\
        \quad $R_{\mathrm{in}}^{(l)} \leftarrow U_{\mathrm{in}} V_{\mathrm{in}}^\top, \quad R_{\mathrm{out}}^{(l)} \leftarrow U_{\mathrm{out}} V_{\mathrm{out}}^\top$
        \STATE Transfer task-vector as in Equation~\ref{eq:full_transfer}: \\
        \quad $\hat{\tau}_B^{(l)} \leftarrow (R_{\mathrm{out}}^{(l)})^\top \tau_A^{(l)} R_{\mathrm{in}}^{(l)}$
    \ENDFOR
    \RETURN ${\theta}_{B,\mathrm{pre}} +\hat\tau_B$
    \end{algorithmic}
\end{algorithm}
\section{Activation consistency along fine-tuning trajectories}
\label{appendix:activation}
BiCo uses pre-trained activations as non-iterative proxies for the input-side factors in the bilinear task-vector decomposition. 
Since the task-vector is formed by accumulating updates along the fine-tuning trajectory, the ideal input-side alignment would reflect the trajectory activations $\vx_t$. 
However, prior work has observed that layer inputs remain relatively stable during fine-tuning~\citep{WUDI}. 
Here, we provide an empirical sanity check of this observation in our transfer setting.

\begin{figure}[ht]
  \centering
  \vspace{-0.6em}
  \includegraphics[width=0.49\linewidth]{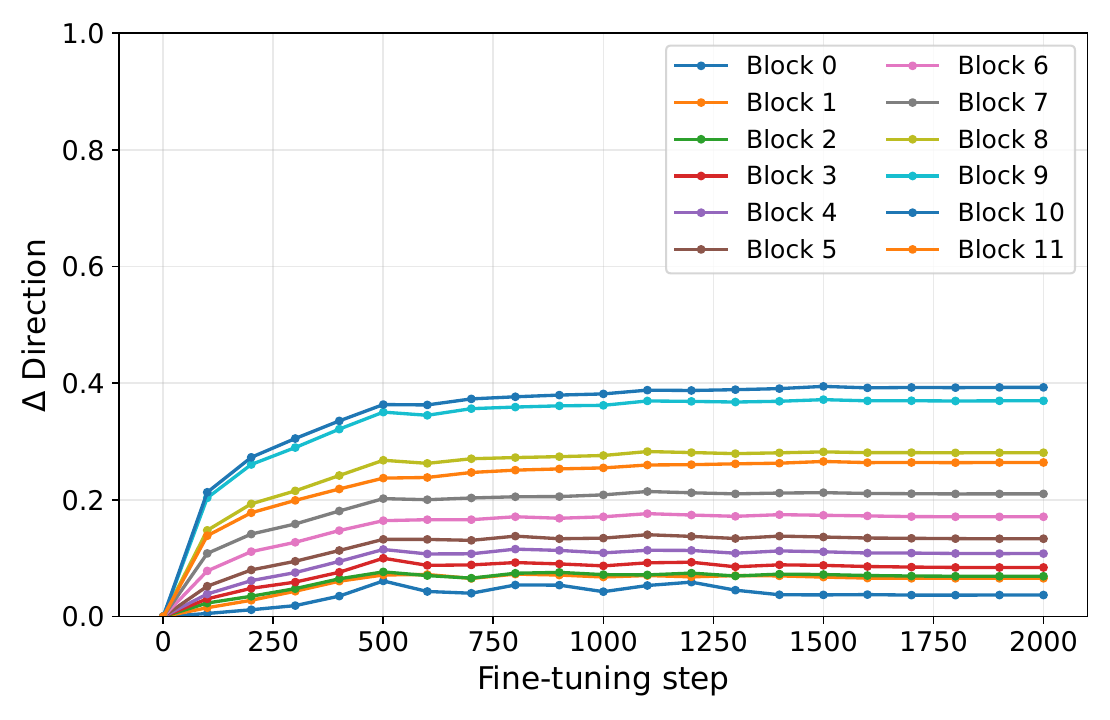}
  \hfill
  \includegraphics[width=0.49\linewidth]{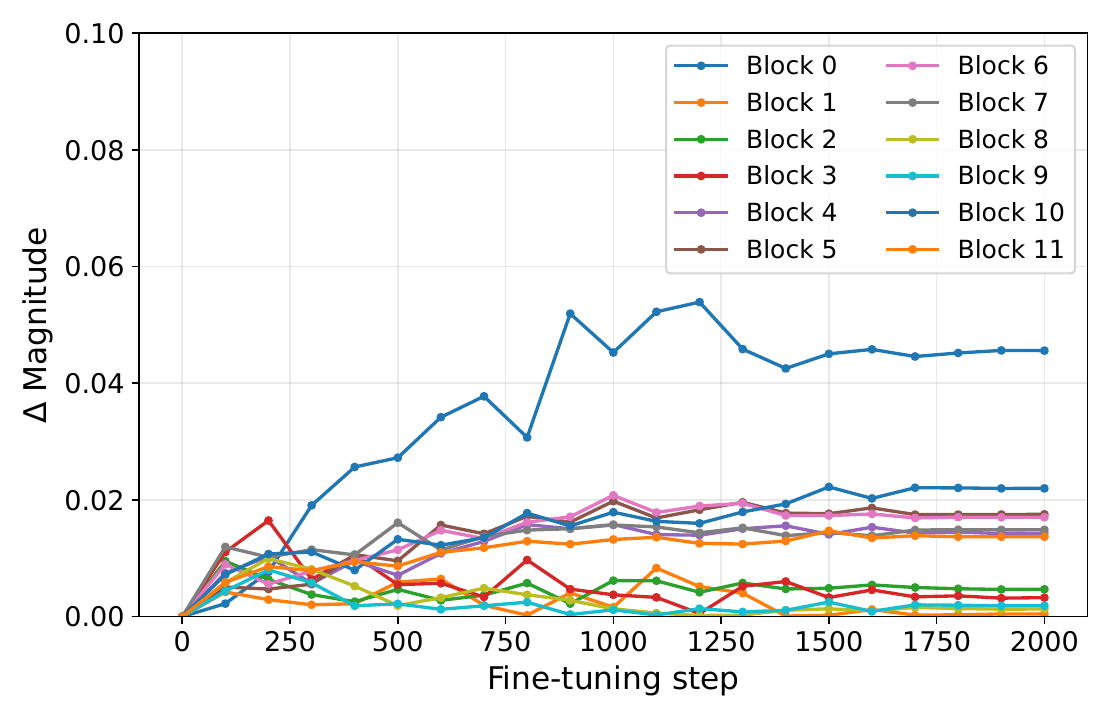}
  \vspace{-0.6em}
  \caption{\textbf{Consistency of pre-trained activations along the fine-tuning trajectory.}
   We measure the consistency between pre-trained activations and activations along the fine-tuning trajectory of ViT-B/16, evaluated every 100 steps and averaged over 8 vision tasks. (Left) directional shift. (Right) magnitude change.}
  \label{fig:step}
  \vspace{-1.5em}
\end{figure}

Following the input-consistency diagnostic of prior work~\citep{WUDI}, we compare the pre-trained activations $\vx_{\mathrm{pre}}$ with activations $\vx_t$ collected along the fine-tuning trajectory. 
For each layer, we measure the directional and magnitude changes as
\begin{equation}
\Delta\text{Direction}
=
1 - \cos(\vx_t, \vx_{\mathrm{pre}}),
\quad
\Delta\text{Magnitude}
=
\frac{
\left|
\left\|\vx_t\right\|_2
-
\left\|\vx_{\mathrm{pre}}\right\|_2
\right|
}{
\left\|\vx_{\mathrm{pre}}\right\|_2
}.
\end{equation}
As shown in Figure~\ref{fig:step}, most layers exhibit high consistency throughout the fine-tuning trajectory: the directional shift typically remains below 0.4, and the magnitude change is also limited across layers. 
This indicates that fine-tuning largely preserves the input-side activation coordinates, rather than substantially changing them at every optimization step. 
These results support using pre-trained activations $\vx_{\mathrm{pre}}$ as practical proxies for the trajectory activation factors $\vx_t$ when estimating the input-side alignment in BiCo.

\section{Cross-width representational similarity analysis}
\label{app:repr_similarity}
In Section~\ref{subsec:inputside}, we use same-width model pairs to analyze both CKA and activation cosine similarity, since raw cosine similarity is only directly defined when the hidden dimensions match.
Here, we further examine whether the shared relational geometry observed in the main text also persists when the source and target models have different hidden widths.

\begin{figure}[ht]
  \centering
  \includegraphics[width=0.60\linewidth]{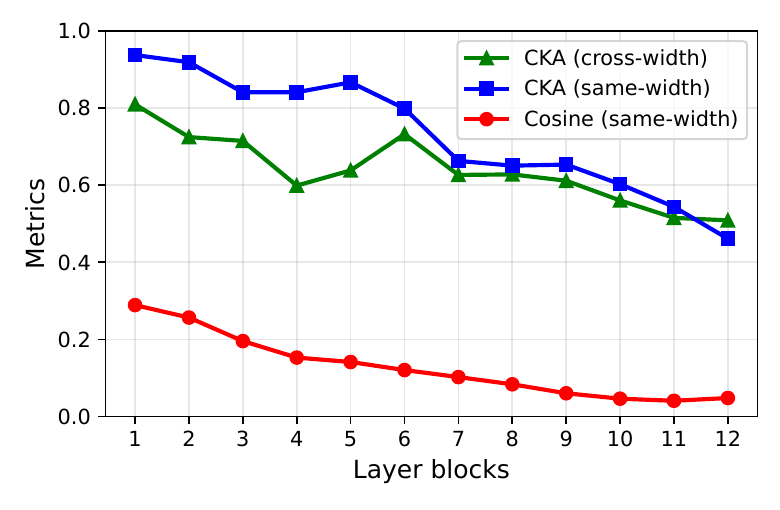}
  \vspace{-0.5em}
  \caption{\textbf{Representational similarity between different pre-trained models.}
   Initial CKA and cosine similarity for same-width (ViT-B/16) and different-width (ViT-B/16-plus) pairs. Results are averaged across 8 vision tasks and layers within each residual block, using $N$=100 calibration samples.}
  \label{fig:appendix1}
  \vspace{-0.5em}
\end{figure}

Figure~\ref{fig:appendix1} reports layer-wise CKA between a ViT-B/16 source model and a wider ViT-B/16-plus target model.
Despite the width mismatch, CKA remains consistently high across corresponding layers, indicating that the models preserve similar sample-wise relational geometry even when their hidden dimensions differ.
This supports our use of feature-space alignment for cross-width task-vector transfer: although raw coordinate-wise comparison is not directly available, the underlying relational structure provides a meaningful basis for estimating alignment maps.

\section{Geometry preservation of orthogonal alignment}
\label{app:orthogonal_preservation}

We justify the geometry-preserving property of the orthogonality constraint used in Equation~\ref{eq:procrustes_x} and Equation~\ref{eq:procrustes_g}, which underlies both our input- and output-side Procrustes alignments.
Let \(\mR \in \mathbb{R}^{d_A \times d_B}\) be an alignment map satisfying
\(\mR\mR^\top = \mI_{d_A}\).
The same argument applies to both the input-side alignment \(\mR_{\mathrm{in}}^{(l)}\) and the output-side alignment \(\mR_{\mathrm{out}}^{(l)}\).

For any source-side vectors
\(\vz, \vz_i, \vz_j \in \mathbb{R}^{1 \times d_A}\),
the mapping \(\vz \mapsto \vz\mR\) preserves norms:
\begin{equation}
\left\|\vz\mR\right\|_2^2
=
\vz\mR\mR^\top \vz^\top
=
\vz\vz^\top
=
\left\|\vz\right\|_2^2,
\end{equation}
and inner products:
\begin{equation}
\left\langle \vz_i\mR, \vz_j\mR \right\rangle
=
\vz_i\mR\mR^\top \vz_j^\top
=
\vz_i\vz_j^\top
=
\left\langle \vz_i, \vz_j \right\rangle .
\end{equation}
Therefore, the alignment preserves source-side norms, inner products, and cosine similarities while mapping source coordinates into the target feature space.
\section{Experimental details}
\label{app:exp}

\subsection{Training details}
Following the fine-tuning protocol established by Task Arithmetic~\citep{ilharco2023editing}, we obtain each task-vector by fine-tuning the pre-trained model for 2,000 training steps  with a batch size of 128 and a learning rate of $1\times10^{-5}$. 
We use the AdamW optimizer with a weight decay of 0.1, together with cosine annealing and 200 warm-up steps.
The dataset has been split into training and validation sets, comprising 90$\%$ and 10$\%$ of the data, respectively.
\paragraph{Compute resources.}
Fine-tuning, performance evaluation, and computational cost measurements for ViT-B/16-scale experiments were conducted on NVIDIA GeForce RTX 3090 GPUs. 
Larger-scale experiments, including those involving ViT-L/14, were conducted on NVIDIA H100 80GB HBM3 GPUs.

\subsection{Dataset details}
We evaluate our method on diverse vision and NLP benchmarks to assess its transferability across modalities and task types. The vision datasets cover remote sensing, traffic sign recognition, digit recognition, texture classification, fine-grained recognition, and scene classification, while the NLP datasets focus on natural language inference and paraphrase identification.
\paragraph{Vision Datasets.} 

\begin{itemize}
\item \textbf{EuroSAT~\citep{eurosat}.} A remote sensing dataset based on Sentinel-2 satellite imagery, consisting of approximately 27,000 images across 10 classes. 
\item \textbf{GTSRB~\citep{gtsrb}.} The German Traffic Sign Recognition Benchmark, a widely used dataset for traffic sign classification, consisting of 51,839 images across 43 categories.
\item \textbf{SVHN~\citep{svhn}.} A real-world digit classification dataset containing house number digits extracted from Google Street View images, with 73,257 images across 10 classes (0–9).
\item \textbf{RESISC45~\citep{resics45}.} A scene classification dataset based on satellite and aerial imagery, comprising 31,500 images across 45 classes.
\item \textbf{DTD~\citep{dtd}.} The Describable Textures Dataset, designed for recognizing and categorizing visual texture patterns, comprising 5,640 images annotated into 47 distinct classes.
\item \textbf{Cars~\citep{cars}.} A fine-grained image classification dataset for distinguishing car models, manufacturers, and years, consisting of 16,185 images across 196 classes.
\item \textbf{MNIST~\citep{mnist}.} The Modified National Institute of Standards and Technology dataset, a benchmark for handwritten digit recognition, composed of grayscale images of digits from 0 to 9 collected from a variety of writers, totaling 70,000 images across 10 classes.
\item \textbf{SUN397~\citep{sun397}.} The SUN397 scene recognition dataset, a large-scale benchmark for scene classification, consisting of 108,754 images across 397 classes.
\end{itemize}

\paragraph{NLP Datasets.} 
\begin{itemize}
\item \textbf{MNLI~\citep{mnli}.} The Multi-Genre Natural Language Inference dataset, a benchmark for natural language inference across diverse text genres, comprising approximately 433,000 training examples of sentence pairs annotated with entailment, contradiction, or neutral relationships.
\item \textbf{QNLI~\citep{GLUE}.} A natural language inference dataset consisting of 104,743 training examples of question–sentence pairs annotated to determine whether the sentence contains the answer to the question.
\item \textbf{RTE~\citep{GLUE}.} The Recognizing Textual Entailment dataset, a benchmark for textual inference, comprising approximately 2,500 training examples of sentence pairs annotated with entailment relationships.
\item \textbf{QQP~\citep{qqp}.} The Quora Question Pairs, a paraphrase identification dataset consisting of approximately 364,000 training examples of question pairs annotated as duplicate or non-duplicate.
\item \textbf{SNLI~\citep{snli}.} The Stanford Natural Language Inference dataset, a large-scale benchmark for natural language inference, comprising approximately 570,000 examples of sentence pairs annotated with entailment, contradiction, or neutral labels.
\end{itemize}
\section{Additional experiments}
\label{app:addexps}

\subsection{Extended results on width scaling}
\paragraph{Robustness to random calibration seeds.}
\begin{table*}[t]
\caption{\textbf{Ablation study on random seed initialization.} Task-vector transfer from a ViT-B/16 model \textit{A} (LAION-2B) to a ViT-B/16-plus model \textit{B} (LAION-400M). Results are averaged over five random seeds. \(N\) denotes the number of calibration samples used to estimate the Procrustes maps. We report mean ± standard deviation for each $N$.}
\label{tab:table11}
\centering
\small
\renewcommand{\arraystretch}{1.05}
\setlength{\tabcolsep}{1.8pt}
\begin{tabular}{lcccccccccc}
\toprule
\textbf{Model}& $N$ & \textbf{EuroSAT} & \textbf{GTSRB} & \textbf{SVHN} & \textbf{RESISC45} & \textbf{DTD} & \textbf{Cars} & \textbf{MNIST} & \textbf{SUN397} & \textbf{AVG} \\
\midrule
$\theta_B$ \textit{zero-shot} & -- & 42.51 & 49.61 & 37.77 & 65.95 & 55.42 & 84.50 & 57.02 & 68.50 & 57.66  \\

\midrule
\mymethod{}&10 & \makecell{50.90 \\ {\scriptsize $\pm$ 6.54}} & \makecell{54.90 \\ {\scriptsize $\pm$ 1.59}}
& \makecell{51.76 \\ {\scriptsize $\pm$ 6.46}} & 
\makecell{66.38 \\ {\scriptsize $\pm$ 1.86}} & 
\makecell{56.06 \\ {\scriptsize $\pm$ 0.87}} & 
\makecell{84.43 \\ {\scriptsize$\pm$0.34}} & 
\makecell{79.55 \\ {\scriptsize$\pm$3.69}} & 
\makecell{68.61 \\ {\scriptsize$\pm$0.10}} & \textbf{\makecell{64.07 \\ {\scriptsize$\pm$ 2.68}}}\\
\midrule
\mymethod{} &20 & \makecell{70.47 \\ {\scriptsize $\pm$ 4.43}} & \makecell{59.64 \\ {\scriptsize $\pm$ 1.49}}
& \makecell{60.66 \\ {\scriptsize $\pm$ 5.71}} & 
\makecell{68.24 \\ {\scriptsize $\pm$ 0.53}} & 
\makecell{56.66 \\ {\scriptsize $\pm$ 1.02}} & 
\makecell{  84.21\\ {\scriptsize$\pm$ 0.59}} &
\makecell{ 86.76\\ {\scriptsize$\pm$2.80}} &
\makecell{ 68.70\\ {\scriptsize$\pm$0.18}} & \textbf{\makecell{ 69.41\\ {\scriptsize$\pm$2.09}}}\\
\midrule
\mymethod{} &50 & \makecell{79.81 \\ {\scriptsize $\pm$ 1.32}} & \makecell{63.84 \\ {\scriptsize $\pm$ 4.16}}
& \makecell{69.52 \\ {\scriptsize $\pm$ 7.24}} & 
\makecell{72.61 \\ {\scriptsize $\pm$ 0.88}} & 
\makecell{58.57 \\ {\scriptsize $\pm$ 0.86}} & 
\makecell{84.72 \\ {\scriptsize$\pm$0.34}} &
\makecell{92.61   \\ {\scriptsize$\pm$1.14}}& 
\makecell{69.13 \\ {\scriptsize$\pm$0.19}}& \textbf{\makecell{73.85 \\ {\scriptsize$\pm$2.02}}}\\
\midrule
\mymethod{} &100 & \makecell{81.74 \\ {\scriptsize $\pm$ 1.62}} & \makecell{67.93 \\ {\scriptsize $\pm$ 2.93}}
& \makecell{74.73 \\ {\scriptsize $\pm$ 4.20}} & 
\makecell{76.81 \\ {\scriptsize $\pm$ 0.55}} & 
\makecell{61.37 \\ {\scriptsize $\pm$ 1.36}} &
\makecell{84.91 \\ {\scriptsize $\pm$0.34}} &
\makecell{93.72 \\ {\scriptsize $\pm$1.35}} &
\makecell{69.41 \\ {\scriptsize$\pm$0.16}}& \textbf{\makecell{76.32 \\ {\scriptsize$\pm$1.56}}}\\
\bottomrule
\end{tabular}
\end{table*}
We conduct an ablation study investigating the sensitivity of \mymethod{} to key factors: the sample size $N\in\{10,20,50,100\}$ and the random seed used for initialization.
Specifically, we consider transferring from a ViT-B/16 model pre-trained on LAION-2B to a ViT-B/16-plus model pre-trained on LAION-400M.
As shown in Table~\ref{tab:table11}, \mymethod{} achieves higher average performance than the zero-shot baseline even with a small number of calibration samples ($N=10$), demonstrating high data efficiency. As the number of samples increases, the average performance improves steadily, while the variance across different random seeds decreases. These findings demonstrate the practical reliability of \mymethod{}, showing that it remains robust to random sampling variation even in data-limited settings.

\paragraph{Sensitivity to calibration set composition.}
\begin{table}[t]
\centering
\caption{
\textbf{Effect of calibration sampling strategy.}
Task-vector transfer from a ViT-B/16 model \textit{A} (LAION-2B) to a ViT-B/16-plus model \textit{B} (LAION-400M).
We compare different ways of constructing the calibration set used to estimate
the Procrustes maps, using \(N=100\) calibration samples for all strategies.
}
\label{tab:table13}
\small
\renewcommand{\arraystretch}{1.05}
\setlength{\tabcolsep}{2.5pt} 
\begin{tabular}{lcccccccccc}
\toprule
\textbf{Strategy} & \textbf{EuroSAT} & \textbf{GTSRB} & \textbf{SVHN} & \textbf{RESISC45} & \textbf{DTD} & \textbf{Cars} & \textbf{MNIST} & \textbf{SUN397} & \textbf{AVG}\\
\midrule
$\theta_B$ \textit{zero-shot} & 42.51 & 49.61 & 37.77 & 65.95 & 55.42 & 84.50 & 57.02 & 68.50 & 57.66  \\
\midrule
\textit{Class-balanced} & 82.48& 68.31&71.78&79.60&62.02&85.11&93.97 & 69.30&  76.57 \\
\textit{Centroid-near} & 78.55& 72.28&71.71&78.92&63.51&85.81 & 93.79& 69.74& 76.78\\
\textit{Centroid-far} & 68.66& 60.97&69.60&72.30&59.89&84.80 &84.52 &69.37 & 71.26 \\
\textit{Half-class} & 63.88& 60.91&72.29&73.93&61.32&85.33 & 75.55& 69.30& 70.31  \\
\textit{One-class} & 53.29& 50.17&63.63&69.50&57.23&84.03 &81.49 &68.88 &  66.02 \\
\midrule
\rowcolor[HTML]{FFF2CC}
\textit{Random} (\mymethod{}) & 80.85& 66.11& 78.63 & 77.49 & 61.86 & 85.16 & 92.93 & 69.31 & 76.54 \\
\bottomrule
\end{tabular}
\end{table}
To isolate the effect of calibration-set composition, we fix the calibration budget to \(N=100\) and vary only the sampling strategy used to construct the calibration set.
We consider six sampling strategies. 
(i) \textit{Random} is our default strategy and samples examples uniformly from the training split. 
(ii) \textit{Class-balanced} samples examples as evenly as possible across classes to improve label coverage. 
(iii) \textit{Centroid-near} computes class centroids in the pre-trained feature space and selects examples closest to their corresponding class centroids. 
(iv) \textit{Centroid-far} uses the same centroids but selects examples farthest from their corresponding class centroids. 
(v) \textit{Half-class} samples examples only from half of the classes. 
(vi) \textit{Single-class} samples all calibration examples from a single class.
Table~\ref{tab:table13} shows that \mymethod{} is largely robust to calibration-set composition. 
Our default \textit{Random} strategy remains competitive with structured alternatives such as \textit{Class-balanced} and
\textit{Centroid-near} sampling, suggesting that carefully curated calibration sets are not necessary.
Performance degrades when calibration samples are drawn from atypical examples or from a restricted set of classes, indicating that
representativeness and label coverage help estimate reliable coordinate maps.
Overall, these results support the use of simple random calibration in our main experiments.

\paragraph{Few-shot fine-tuning under matched computational cost.}
\begin{table}[t]
\centering
\caption{
\textbf{Width scaling under matched computational cost.}
Task-vector transfer from a ViT-B/16 model A (LAION-2B) to a wider ViT-B/16-plus model B (LAION-400M).
$K$ denotes the number of calibration examples per class used to estimate the Procrustes maps and to train the few-shot fine-tuning baselines.
$\theta_B^{\mathrm{1st}}$ denotes target models fine-tuned on the same $K$-shot calibration set for 1 step.
$\Delta$Acc denotes the accuracy gain over the corresponding $\theta_B^{\mathrm{1st}}$ baseline under the same $K$-shot setting.
}
\label{tab:table10}
\small
\renewcommand{\arraystretch}{1.05}
\setlength{\tabcolsep}{2.5pt} 
\begin{tabular}{lccccccccccc}
\toprule
\textbf{Model} & $K$ &  \textbf{EuroSAT} & \textbf{GTSRB} & \textbf{SVHN} & \textbf{RESISC45} & \textbf{DTD} & \textbf{Cars} & \textbf{MNIST} & \textbf{SUN397} & \textbf{AVG ($\Delta$Acc)}\\
\midrule
$\theta_B$ \textit{zero-shot} & -- & 42.51 & 49.61 & 37.77 & 65.95 & 55.42 & 84.50 & 57.02 & 68.50 & 57.66 {\color{black}(-)} \\
$\theta_B$ \textit{fine-tune} & -- & 99.00 & 99.19 & 97.80 & 96.46 & 82.55 & 90.11 & 99.81 & 78.29 & 92.90 {\color{black}(-)} \\
$\theta_A$ \textit{fine-tune} & -- & 94.67 & 99.03 & 97.89 & 96.81 & 83.46 & 92.10 & 99.72 & 79.35 & 92.88 {\color{black}(-)} \\
$\theta_B + \tau_A^{\text{padded}}$ & -- & 46.41 & 38.21 & 27.27 & 62.57 & 53.56 & 82.00 & 51.70 & 68.29 & 53.75 {\color{black}(-)} \\
\midrule
$\theta_B^{\mathrm{1st}}$  & 1 & 60.96&	56.47&	41.09&	68.68&	56.22&	78.67&	81.76&	69.11&	64.12 \color{black}(+0.00)\\
THESEUS & 1 &44.55& 53.79&45.87&68.04&56.11&84.46&57.01& 69.10& 59.86 \color{red!80!black}(-4.26) \\
\rowcolor[HTML]{FFF2CC}
\textbf{\mymethod} & 1 & 66.88& 62.26&65.41&73.20& 59.20& 85.46& 85.95 &70.40 &\textbf{71.09} \textbf{{\color{green!70!black}(+6.97)}} \\
\midrule
$\theta_B^{\mathrm{1st}}$  & 2 & 62.48&	58.77&	56.95&	71.06&	56.28&	77.01&	83.74&	69.17&	66.93 \color{black}(+0.00) \\
THESEUS & 2 & 44.55&53.99&48.38&68.33&56.27&84.50&55.48&69.08& 60.07 \color{red!80!black}(-6.86) \\
\rowcolor[HTML]{FFF2CC}
\textbf{\mymethod} & 2 & 71.33& 67.91&73.92&77.84&62.28& 86.58& 91.87& 70.57 &\textbf{75.28} \textbf{{\color{green!70!black}(+8.35)}} \\
\midrule
$\theta_B^{\mathrm{1st}}$  & 5 & 62.52&	59.83&	55.49&	72.83&	56.6&	75.7&	85.03&	69.34&	67.16 \color{black}(+0.00)\\
THESEUS & 5 & 49.66&55.01&50.59&68.88&56.70&84.69&59.25&68.91&61.71 {\color{red!80!black}(-5.45)} \\
\rowcolor[HTML]{FFF2CC}
\textbf{\mymethod} & 5 & 79.25& 74.05&77.51&81.85&64.25& 86.94& 93.50& 70.96& \textbf{78.53} \textbf{{\color{green!70!black}(+11.37)}} \\
\bottomrule
\end{tabular}
\end{table}
We further compare \mymethod{} with a few-shot target fine-tuning baseline using the same calibration data budget and a matched forward--backward computational budget.
Table~\ref{tab:table10} reports the results, where $K$ denotes the number of calibration samples per class.
$\theta_B^{\mathrm{1st}}$ denotes the target model fine-tuned on the same $K$-shot calibration set for one step, matching the forward--backward budget used by \mymethod{} to estimate the alignment maps.
Compared with $\theta_B^{\mathrm{1st}}$, which directly updates the target parameters under the matched budget, \mymethod{} achieves substantially higher performance across different values of $K$.
This result suggests that using the calibration set to estimate bilinear coordinate alignment provides a more effective use of limited supervision than directly applying a small number of gradient updates to the target model.
Importantly, \mymethod{} obtains these gains without iterative target-model fine-tuning, highlighting its practical utility as a training-free alternative to few-shot target adaptation.

\paragraph{Transfer from wide to narrow models.}
\begin{table*}[t]
\caption{\textbf{Reverse width scaling (wide → narrow).}
Task-vector transfer from a ViT-B/16-plus model \textit{A} (LAION-400M) to a narrower ViT-B/16 model \textit{B} (LAION-2B). \(N\) denotes the number of calibration samples used to estimate the Procrustes maps. $\Delta$Acc denotes accuracy gain over model B zero-shot baseline.}
\label{tab:table6}
\centering
\small
\renewcommand{\arraystretch}{1.05}
\setlength{\tabcolsep}{1.8pt}
\begin{tabular}{lcccccccccc}
\toprule
\textbf{Model} & $N$ & \textbf{EuroSAT} & \textbf{GTSRB} & \textbf{SVHN} & \textbf{RESISC45} & \textbf{DTD} & \textbf{Cars} & \textbf{MNIST} & \textbf{SUN397} & \textbf{AVG ($\Delta$Acc)} \\
\midrule
$\theta_B$ \textit{zero-shot} & -- & 50.15 & 48.35 & 50.05 & 68.22 & 55.96 & 88.56 & 65.79 & 70.44 & 62.19 {\color{black}(+0.00)}  \\
$\theta_B$ \textit{fine-tune} & -- & 94.67 & 99.03 & 97.89 & 96.81 & 83.46 & 92.10 & 99.72 & 79.35 & 92.88 {\color{green!70!black}(+30.69)} \\
$\theta_A$ \textit{fine-tune} & -- &  99.00 & 99.19 & 97.80 & 96.46 & 82.55 & 90.11 & 99.81 & 78.29 & 92.90 {\color{green!70!black}(+30.71)} \\
$\theta_B + \tau_A^{\text{crop}}$ & -- &  33.11&	36.18&	19.73&	45.10&	55.53	&74.13	&36.79	&57.18&44.72 {\color{red!80!black}(-17.47)} \\
\midrule
THESEUS  & 10 &50.07&	54.94&	61.70&	68.52&	55.59&	87.60&	71.86&	70.51&  65.10 {\color{green!70!black}(+2.91)} \\
\rowcolor[HTML]{FFF2CC}
\textbf{\mymethod} & 10 & 20.37&	56.74&	68.59&	68.43&	57.61&	85.60&	88.14&	70.07& \textbf{64.44} \textbf{{\color{green!70!black}(+2.25)}}\\
\midrule
THESEUS & 20 &49.41&	57.40&	65.25&	70.48&	56.28&	88.05&	73.54&	70.14& 66.32  {\color{green!70!black}(+4.13)}\\
\rowcolor[HTML]{FFF2CC}
\textbf{\mymethod} & 20 &59.00&	67.07&	80.43&	71.11&	58.24&	86.05&	70.09&	70.50& \textbf{70.31} \textbf{{\color{green!70!black}(+8.12)}} \\
\midrule
THESEUS & 50 & 51.56&	59.47&	65.95&	72.25& 56.12&	87.97&	79.55&	70.21& 67.89 {\color{green!70!black}(+5.70)}\\
\rowcolor[HTML]{FFF2CC}
\textbf{\mymethod} & 50 & 77.15&	72.60&	87.12&	74.90&	62.02&	86.27&	87.15&	70.80& \textbf{77.25} \textbf{{\color{green!70!black}(+15.06)}} \\
\midrule
THESEUS & 100 &51.85&	60.97&	69.22&	73.10&	57.82&	87.86&	78.54&	70.34& 68.71 {\color{green!70!black}(+6.52)} \\
\rowcolor[HTML]{FFF2CC}
\textbf{\mymethod} & 100 & 88.33&	78.39&	89.10&	79.32&	65.05&	86.79&	93.19&	71.21& \textbf{81.42} \textbf{{\color{green!70!black}(+19.23)}} \\
\bottomrule
\end{tabular}
\end{table*}
We further evaluate whether task-vector transfer remains effective when transferring from a wide model to a narrow model. Specifically, we consider transferring the task-vector $\tau_A$ obtained from a ViT-B/16-plus model \textit{A} pre-trained on LAION-400M to a ViT-B/16 target model \textit{B} pre-trained on LAION-2B.
As shown in Table~\ref{tab:table6}, an approach that resolves the dimensional mismatch by simply cropping the task-vector ($\theta_B + \tau_A^{\text{crop}}$) results in performance that falls even below the zero-shot baseline of the target model. This indicates that there exists an intrinsic mismatch between the parameter spaces of structurally different models, which cannot be eliminated by straightforward dimensionality reduction, and suggests that reverse-direction transfer is infeasible without explicit alignment. 
In contrast, \mymethod{} consistently outperforms the zero-shot baseline with only a small number of calibration samples, and steadily approaches the fine-tuned performance as the number of samples $N$ increases. This demonstrates that our proposed method operates robustly regardless of the transfer direction.

\paragraph{Generalization to different pre-training sources.}
\begin{table*}[t]
\caption{\textbf{Width scaling (narrow → wide) under different pre-training sources.}
Task-vector transfer from a ViT-B/16 model \textit{A} (Datacomp-XL) to a wider ViT-B/16-plus model \textit{B} (LAION-400M). \(N\) denotes the number of calibration samples used to estimate the Procrustes maps. $\Delta$Acc denotes accuracy gain over model B zero-shot baseline.
}
\label{tab:table7}
\centering
\small
\renewcommand{\arraystretch}{1.05}
\setlength{\tabcolsep}{1.8pt}

\begin{tabular}{lcccccccccc}
\toprule
\textbf{Model} & $N$ & \textbf{EuroSAT} & \textbf{GTSRB} & \textbf{SVHN} & \textbf{RESISC45} & \textbf{DTD} & \textbf{Cars} & \textbf{MNIST} & \textbf{SUN397} & \textbf{AVG ($\Delta$Acc)} \\
\midrule
$\theta_B$ \textit{zero-shot} & -- & 42.51 & 49.61 & 37.77 & 65.95 & 55.42 & 84.50 & 57.02 & 68.50 & 57.66 {\color{black}(+0.00)} \\
$\theta_B$ \textit{fine-tune} & -- & 99.00 & 99.19 & 97.80 & 96.46 & 82.55 & 90.11 & 99.81 & 78.29 & 92.90 {\color{green!70!black}(+35.24)} \\
$\theta_A$ \textit{fine-tune} & -- & 98.96 & 99.06 & 98.02 & 96.71 & 84.25 & 92.50 & 99.77 & 79.25 & 93.57 {\color{green!70!black}(+35.91)} \\
$\theta_B + \tau_A^{\text{padded}}$ & -- & 38.07 & 44.17 & 31.03 & 60.95 & 54.79 & 82.83 & 15.64 & 68.00 & 49.44 {\color{red!80!black}(-8.22)} \\
\midrule
THESEUS & 10 & 42.22 & 51.72 & 38.71 & 67.38 & 55.53 & 84.11 & 56.94 & 68.55 & 58.15 {\color{green!70!black}(+0.49)} \\
\rowcolor[HTML]{FFF2CC}
\textbf{\mymethod} & 10 & 46.48 & 54.40 & 31.90 & 66.08 & 57.02 & 84.59 & 75.08 & 68.58 & \textbf{60.52} \textbf{{\color{green!70!black}(+2.86)}} \\
\midrule
THESEUS & 20 & 44.48 & 52.51 & 41.20 & 68.17 & 56.17 & 84.16 & 57.43 & 68.44 & 59.07 {\color{green!70!black}(+1.41)} \\
\rowcolor[HTML]{FFF2CC}
\textbf{\mymethod} & 20 & 66.93 & 56.67 & 46.88 & 68.03 & 58.35 & 84.60 & 81.10 & 68.81 & \textbf{66.42} \textbf{{\color{green!70!black}(+8.76)}} \\
\midrule
THESEUS & 50 & 47.37 & 52.82 & 41.80 & 68.75 & 55.74 & 84.40 & 61.93 & 68.73 & 60.19 {\color{green!70!black}(+2.53)} \\
\rowcolor[HTML]{FFF2CC}
\textbf{\mymethod} & 50 & 80.22 & 58.59 & 66.69 & 73.46 & 60.43 & 84.90 & 90.21 & 69.09 & \textbf{72.95} \textbf{{\color{green!70!black}(+15.29)}} \\
\midrule
THESEUS & 100 & 48.15 & 52.83 & 43.42 & 69.11 & 55.85 & 84.39 & 61.67 & 68.83 & 60.53 {\color{green!70!black}(+2.87)} \\
\rowcolor[HTML]{FFF2CC}
\textbf{\mymethod} & 100 & 83.41 & 65.92 & 73.28 & 78.05 & 62.66 & 85.23 & 90.16 & 69.46 & \textbf{76.02} \textbf{{\color{green!70!black}(+18.36)}} \\
\bottomrule
\end{tabular}

\end{table*}
We further analyze task-vector transfer when the source and target models are pre-trained on different data sources, while retaining the narrow-to-wide transfer setting.
In this setup, the source model \textit{A} is a ViT-B/16 pre-trained on Datacomp-XL, whereas the target model \textit{B} is a
ViT-B/16-plus pre-trained on LAION-400M, introducing pre-training mismatch in addition to architectural discrepancy.
As shown in Table~\ref{tab:table7}, transfer via zero-padding again results in a significant performance drop, indicating that discrepancies in both model width and pre-training data distributions further aggravate the misalignment between parameter spaces. \mymethod{} achieves robust transfer performance even in this challenging scenario. With only a small number of calibration samples, it consistently surpasses the zero-shot baseline, and its performance improves steadily as $N$ increases. Moreover, it significantly outperforms THESEUS across all values of $N$, demonstrating that our bilinear alignment effectively handles both architectural and data-induced discrepancies.

\subsection{Extended results on identical architectures}
\paragraph{Scalability to few-shot $K$.}
\begin{table}[t]
\caption{\textbf{Identical-architecture transfer with varying numbers of shots $K$.} Task-vector transfer is performed between ViT-B/16 models with the same architecture but different pretraining (A: Datacomp-XL → B: LAION-2B). $K$ denotes the number of calibration examples per class used to estimate the Procrustes maps and to train the few-shot fine-tuning baselines. $\Delta$Acc denotes accuracy gain over model B zero-shot baseline.}
\label{tab:table8}
\centering
\small
\renewcommand{\arraystretch}{1.05}
\setlength{\tabcolsep}{1.8pt} 
\begin{tabular}{lccccccccccc}
\toprule
\textbf{Model} & $K$ &  \textbf{EuroSAT} & \textbf{GTSRB} & \textbf{SVHN} & \textbf{RESISC45} & \textbf{DTD} & \textbf{Cars} & \textbf{MNIST} & \textbf{SUN397} & \textbf{AVG ($\Delta$Acc)}\\
\midrule
$\theta_B$ \textit{zero-shot} & -- & 50.15 & 48.35 & 50.05 & 68.22 & 55.96 & 88.56 & 65.79 & 70.44 & 62.19 {\color{black}(+0.00)} \\
$\theta_B$ \textit{fine-tune} & -- & 94.67 & 99.03 & 97.89 & 96.81 & 83.46 & 92.10 & 99.72 & 79.35 & 92.88 {\color{green!70!black}(+30.69)} \\
$\theta_B + \tau_A$ & -- & 36.07 & 33.96 & 20.33 & 63.16 & 54.73 & 86.82 & 72.09 & 66.74 & 54.24 {\color{red!70!black}(-7.95)} \\
TransFusion & -- &  50.30 & 49.48 &52.57 & 68.27 & 56.65 & 88.60 & 73.51 & 70.51 & 63.74 {\color{green!70!black}(+1.55)} \\
\midrule
\rowcolor[HTML]{FFF2CC}
\textbf{\mymethod} & 1 & 68.00 & 74.47 & 74.94 & 71.98 & 63.24 & 87.76 & 91.48 & 72.21 & \textbf{75.51} \textbf{{\color{green!70!black}(+13.32)}} \\
\midrule
\rowcolor[HTML]{FFF2CC}
\textbf{\mymethod} & 2 & 73.96 & 79.50 & 71.32 & 78.60 & 67.07 & 87.53 & 91.39 & 72.44 & \textbf{77.73} \textbf{{\color{green!70!black}(+15.54)}} \\
\midrule
\rowcolor[HTML]{FFF2CC}
\textbf{\mymethod} & 5 & 78.25 & 81.96 & 72.69 & 85.65 & 68.45 & 87.68 & 92.21 & 73.11 & \textbf{80.00} \textbf{{\color{green!70!black}(+17.81)}}\\
\midrule
GradFix & 10 & 64.33 & 65.32 & 66.52 & 73.07 & 59.57 & 81.45 & 89.10 & 70.98 & 71.29 {\color{green!70!black}(+9.10)} \\
THESEUS & 10 & 60.51 & 64.55 & 63.68 & 72.57 & 58.24 & 87.29 & 87.56 & 70.82 & 70.65 {\color{green!70!black}(+8.46)} \\
\rowcolor[HTML]{FFF2CC}
\textbf{\mymethod} & 10 & 79.92 & 84.22 & 80.87 & 86.47 & 72.87 & 88.14 & 92.36 & 73.29 & \textbf{82.27} \textbf{{\color{green!70!black}(+20.08)}} \\
\midrule
GradFix & 20 & 64.56 & 64.56 & 66.60 & 73.54 & 59.84 & 81.82 & 88.62 & 70.83 & 71.30 {\color{green!70!black}(+9.11)} \\
THESEUS & 20 & 56.48 & 64.72 & 68.34 & 72.50 & 59.20 & 87.27 & 89.18 & 70.77 & 71.06 {\color{green!70!black}(+8.87)} \\
\rowcolor[HTML]{FFF2CC}
\textbf{\mymethod} & 20 & 82.03 & 83.11 & 82.73 & 86.52 & 74.20 & 88.37 & 93.60 & 73.38 & \textbf{82.99} \textbf{{\color{green!70!black}(+20.80)}} \\
\bottomrule
\end{tabular}
\end{table}
We extend the identical-architecture setting by evaluating \mymethod{} under larger data regimes. Table~\ref{tab:table8} presents the full results with increased shot numbers $K\in\{10,20\}$.
The results show that the performance trends observed in the low-data regime persist as the number of calibration samples increases.
In particular, \mymethod{} continues to demonstrate strong data efficiency and consistent improvements over all baselines. As $K$ grows, our method exhibits a stable scaling behavior, further improving performance and narrowing the gap toward full fine-tuning.
Moreover, the performance margin over competing methods such as THESEUS and GradFix remains substantial even at higher data budgets, indicating that our bilinear alignment not only performs well in few-shot settings but also scales effectively with additional data.

\subsection{Extended results on width and depth scaling}
\label{extend}
\paragraph{Transfer from deeper to shallower models.}
\begin{table}[t]
\caption{\textbf{Deeper to shallower model.}
Task-vector transfer from a  ViT-L/14 model to ViT-B/16 model, both pretrained on Datacomp-XL. \(N\) denotes the number of calibration samples used to estimate the Procrustes maps. $\Delta$Acc denotes accuracy gain over model B zero-shot baseline.}
\centering
\small
\renewcommand{\arraystretch}{1.05}
\setlength{\tabcolsep}{1.8pt} 
\begin{tabular}{lcccccccccc}
\toprule
\textbf{Model} & $N$ & \textbf{EuroSAT} &  \textbf{GTSRB} & \textbf{SVHN} & \textbf{RESISC45} & \textbf{DTD} & \textbf{Cars} & \textbf{MNIST} & \textbf{SUN397} & \textbf{AVG ($\Delta$Acc)} \\
\midrule
$\theta_B$ \textit{zero-shot} & -- &  45.81 &55.44&	62.71& 68.60&58.03&	88.77&76.43	& 70.03	&65.72  {\color{black}(+0.00)}  \\
$\theta_B$ \textit{fine-tune} & -- & 98.96&	99.06&	98.02&	96.71&	84.25&	92.50&	99.77&	79.25& 93.57 {\color{green!70!black}(+27.85)} \\
$\theta_A$ \textit{fine-tune} & -- & 99.33&	99.29&	98.21&	97.58&	86.11&	95.22&	99.74&	82.40 &94.74 {\color{green!70!black}(+29.02)}\\
$\theta_B + \tau_A^{\text{crop}}$ & -- & 43.33&54.71 & 63.63&68.44&57.97 & 88.83 & 75.91&69.71 & 65.31 {\color{red!80!black}(-0.41)} \\
\midrule
THESEUS & 100 & 52.66 &	62.55&	71.05&	74.33&	67.81& 88.58 &	82.96&	69.97& 71.23 {\color{green!70!black}(+5.51)}\\
\rowcolor[HTML]{FFF2CC}
\textbf{\mymethod} & 100 & 83.85& 79.80 & 79.85 & 76.55 & 63.29 & 88.33 & 97.44 & 70.12 & \textbf{79.90} \textbf{{\color{green!70!black}(+14.18)}}\\
\toprule
\end{tabular} 
\label{tab:table12}
\end{table}

We further examine a more challenging transfer scenario in which the source model is deeper and wider than the target model. 
Specifically, we transfer task vectors from ViT-L/14 to ViT-B/16, both pre-trained on the Datacomp-XL distribution, thereby introducing mismatches in both depth and hidden width. 
As shown in Table~\ref{tab:table12}, despite this structural discrepancy, \mymethod{} maintains strong transfer performance, indicating that the learned task-specific updates can be effectively translated into the coordinate system of a smaller target model. 
This result highlights the effectiveness of our depth-matching strategy, which provides reliable layer correspondences when transferring across models with different numbers of transformer blocks.

\paragraph{Transfer under different pre-training distributions.}
To further assess the robustness of our approach, we consider a more challenging setting where the source and target models are pre-trained on different data distributions. Specifically, we transfer task vectors from a ViT-B/16 model trained on LAION-2B to a ViT-L/14 model trained on Datacomp-XL.
Unlike the previous setting where both models share the same pre-training distribution, this setting introduces a distribution shift in addition to architectural differences. We apply the same layer matching and Procrustes-based transfer procedure without any modification. As shown in Table~\ref{tab:table9}, our method remains effective under this cross-distribution transfer setting. Despite the mismatch in both model structure and pre-training data, our approach consistently improves over the target zero-shot baseline by a significant margin. This suggests that the proposed transfer mechanism captures transferable task information that generalizes beyond both architectural and data distribution differences.
\begin{table}[t]
\caption{\textbf{Depth and width scaling across different pre-training distributions.}
Task-vector transfer from a  ViT-B/16 model \textit{A} (LAION-2B) to ViT-L/14 model \textit{B} (Datacomp-XL). \(N\) denotes the number of calibration samples used to estimate the Procrustes maps. $\Delta$Acc denotes accuracy gain over model B zero-shot baseline.}
\label{tab:table9}
\centering
\small
\renewcommand{\arraystretch}{1.05}
\setlength{\tabcolsep}{1.8pt} 
\begin{tabular}{lcccccccccc}
\toprule
\textbf{Model} & $N$ & \textbf{EuroSAT} &  \textbf{GTSRB} & \textbf{SVHN} & \textbf{RESISC45} & \textbf{DTD} & \textbf{Cars} & \textbf{MNIST} & \textbf{SUN397} & \textbf{AVG ($\Delta$Acc)} \\
\midrule
$\theta_B$ \textit{zero-shot} & -- &  64.11& 58.91&	67.67&	72.60&	66.96&	93.06&	86.63&	73.77& 70.29 {\color{black}(+0.00)}  \\
$\theta_B$ \textit{fine-tune} & -- & 99.33&	99.30&	98.21&	97.59&	86.12&	95.22&	99.74&	82.40 &94.74 {\color{green!70!black}(+24.45)}\\
$\theta_A$ \textit{fine-tune} & -- & 94.67&	99.03&	97.89&	96.81&	83.46&	92.10&	99.72&	79.35& 92.88 {\color{green!70!black}(+22.59)} \\
$\theta_B + \tau_A^{\text{padded}}$ & -- & 64.03& 58.93&67.88& 72.53& 66.75&93.13&86.56&73.93& 72.96 {\color{green!70!black}(+2.67)} \\
\midrule
THESEUS & 100 & 66.52&	58.93&	68.40&	68.79&	58.24&	93.17&	87.63&	73.97&73.88 {\color{green!70!black}(+3.59)}\\
\rowcolor[HTML]{FFF2CC}
\textbf{\mymethod} & 100 & 80.19&	82.53&	83.40&	83.13 & 71.54 & 93.79 & 98.29 & 74.74 & \textbf{83.45} \textbf{{\color{green!70!black}(+13.16)}}\\
\toprule
\end{tabular} %
\end{table}

\section{Computational cost analysis}
\label{app:computation}
Following standard scaling-law approximations~\citep{cost}, we adopt commonly used per-parameter FLOP
estimates: one forward pass costs approximately $2P$ FLOPs, one backward pass costs
approximately $4P$ FLOPs, and one Adam/AdamW optimizer update costs approximately
$10P$ FLOPs. Let $P_A$ and $P_B$ denote the number of parameters in the source and target
models, respectively. We count one forward/backward pass as one complete pass over the
calibration set.

\mymethod{} assumes that the source task vector is already available, as in the standard task-vector
transfer setting. To construct the transferred target task vector, \mymethod{} performs one
forward--backward pass through both frozen pretrained models to collect input-side activations
and output-side gradients. The resulting calibration cost is approximately $6P_A + 6P_B$.
\mymethod{} additionally performs layer-wise Procrustes estimation and applies the bilinear
task-vector transformation in Equation~\ref{eq:full_transfer}, which we denote by $C_{\mathrm{align}}$.
Therefore, the total cost of \mymethod{} is
\begin{equation}
\label{eq:bico-compute-cost}
\mathrm{Cost}_{\mathrm{BiCo}}
\approx
6P_A + 6P_B + C_{\mathrm{align}} .
\end{equation}
Importantly, \mymethod{} performs zero optimizer updates and requires no task-specific
hyperparameter search.
By contrast, direct target fine-tuning with Adam/AdamW for 2000 optimization steps, following
our fine-tuning protocol, costs
\begin{equation}
\label{eq:ft-compute-cost}
\mathrm{Cost}_{\mathrm{FT}}
\approx
2000 \times (2P_B + 4P_B + 10P_B)
=
32000P_B .
\end{equation}
For comparable-size source and target models, i.e., $P_A \approx P_B = P$, the calibration
forward--backward component of \mymethod{} costs approximately $12P$, whereas full target
fine-tuning costs $32000P$. This highlights that the dominant difference comes from
the absence of repeated optimizer updates in \mymethod{}.

\paragraph{Further analysis of alignment cost.}
We further decompose the one-time overhead $C_{\mathrm{align}}$ under the representative
width-expansion setting, where the source width is $d_A$ and the target width is $d_B$ with
$d_A \le d_B$. Here, $d_A$ and $d_B$ denote representative widths, or equivalently upper
bounds on the input and output dimensions of the transferred matrices. Although individual
matrices may have different shapes, such as attention projections and MLP projections, these
dimensions are constant-factor multiples of the model width in the Transformer architectures
considered in our experiments. Thus, we use $d_A$ and $d_B$ for a simplified asymptotic
analysis; the exact cost is obtained by summing the same terms over all transferred matrices.

Let $L$ denote the number of transferred weight matrices and $M$ the number of calibration
tokens. For each transferred matrix, \mymethod{} estimates two Procrustes maps: one for the
input-side activations and one for the output-side gradients. Each map requires forming a
cross-covariance matrix, whose cost is upper bounded by $\mathcal{O}(M d_A d_B)$.
Computing the SVD of the resulting $d_A \times d_B$ matrix and forming the
Procrustes map both have cost $\mathcal{O}(d_A^2 d_B)$ under $d_A \le d_B$.
Therefore, the Procrustes estimation cost across all transferred matrices is
\begin{equation}
\label{eq:procrustes-cost}
C_{\mathrm{proc}}
=
\mathcal{O}\!\left(
2L(M d_A d_B + d_A^2 d_B)
\right).
\end{equation}
where the factor of $2$ accounts for the input- and output-side alignments.

After estimating the Procrustes maps, \mymethod{} applies them to each source task-vector
matrix as in Equation~\ref{eq:full_transfer}.
Using the same representative dimensions, this requires two dense matrix multiplications,
with cost upper bounded by
\begin{equation}
\label{eq:apply-cost}
C_{\mathrm{apply}}
=
\mathcal{O}\!\left(
L(d_A^2 d_B + d_A d_B^2)
\right).
\end{equation}
Thus, the one-time alignment overhead is
\begin{equation}
\label{eq:alignment-cost}
C_{\mathrm{align}}
=
C_{\mathrm{proc}} + C_{\mathrm{apply}} .
\end{equation}
When the source and target widths are comparable, i.e., $d_A \approx d_B \approx d$, the
above cost simplifies, up to constant factors, to
\begin{equation}
\label{eq:alignment-cost-simplified}
C_{\mathrm{align}}
=
\mathcal{O}\!\left(L(Md^2+d^3)\right).
\end{equation}
\section{Limitations}
\label{app:limitations}
\mymethod{} assumes that the source and target models have sufficiently compatible internal structures, such that their layer-wise activation and gradient spaces can be related through orthogonal transformations.
This assumption is reasonable in many practical model-update scenarios, where new model variants often preserve the same high-level backbone while changing scale, pre-training data, or training recipes, as in common ViT and T5 model families.
Accordingly, our experiments focus on transferring between models within related backbone families, including differences in width, depth, and pre-training configuration.
However, transferring between substantially different architectures, tokenization schemes, or modality-specific processing pipelines may require more general alignment mechanisms.
Extending task-vector transfer to such heterogeneous model families is an important direction for future research.

\mymethod{} also requires a small calibration set to estimate the Procrustes mappings and compute output-side gradients.
Although the quality and representativeness of the calibration data can affect transfer performance, this requirement is substantially weaker than target-model fine-tuning, since \mymethod{} does not update target parameters and uses the calibration set only to estimate coordinate correspondences.
Moreover, our results show that even small calibration budgets can yield consistent improvements over zero-shot target models and strong gains over existing transfer baselines, suggesting that the proposed alignment direction is effective in data-limited settings.
Future work may further improve robustness by developing principled calibration-sample selection strategies or data-free approximations for estimating alignment maps.

Our current depth-mismatch setting relies on a simple layer-index matching rule.
While this rule may not be optimal for highly heterogeneous architectures, it is intentionally lightweight and already provides substantial improvements in our evaluated depth-mismatch settings.
This suggests that even coarse structural correspondences can be sufficient for effective task-vector transport when the underlying models share related representational hierarchies.
More adaptive layer matching, such as similarity-based or many-to-one layer correspondence, could further improve performance and broaden applicability.

Finally, although \mymethod{} narrows the gap to direct target fine-tuning, it does not fully eliminate it.
This indicates that some task-specific information may not be captured by linear coordinate alignment alone.
Nevertheless, under the same transfer setting, \mymethod{} substantially improves over prior approaches and consistently outperforms the target zero-shot baseline, demonstrating the promise of bilinear coordinate alignment as a training-free alternative to repeated fine-tuning.
Closing the remaining gap to fully fine-tuned target models remains an important future direction.

\section{Broader impacts}
\label{app:borader}
This work aims to improve the reuse of fine-tuned expertise across evolving pre-trained model families. 
By transferring task vectors without additional target-model fine-tuning, \mymethod{} may reduce the computational cost, storage
overhead, and energy consumption associated with repeatedly adapting newly released model variants.
This can make model adaptation more accessible to researchers and practitioners with limited computational resources.
At the same time, training-free transfer does not by itself correct undesirable properties already encoded in the source task vector, the source fine-tuned model, or the calibration data.
As a result, biases, spurious correlations, failure modes, or unsafe task-specific behaviors may be propagated to the target
model. 
Therefore, transferred models should be evaluated with the same care as directly fine-tuned models, especially before deployment in sensitive or high-stakes domains. 
Appropriate auditing, calibration-data screening, and downstream robustness and safety evaluation remain necessary when applying
task-vector transfer in practice.

\end{document}